\documentclass[a4paper,fleqn]{cas-sc}

\usepackage[authoryear]{natbib}
\usepackage{placeins}
\usepackage{comment}
\def\tsc#1{\csdef{#1}{\textsc{\lowercase{#1}}\xspace}}
\tsc{WGM}
\tsc{QE}
\tsc{EP}
\tsc{PMS}
\tsc{BEC}
\tsc{DE}

\newcounter{TaliaCount}
\addtocounter{TaliaCount}{1}

\begin{document}
\let\WriteBookmarks\relax

\shorttitle{Model for Woven Systems}
\shortauthors{Pradhan et al.} 


\newcommand{\Talia}[1]{{\normalsize{\textbf{({\color{blue}Talia:\ }#1)}}}}
\newcommand{\Anvay}[1]{{\normalsize{\textbf{({\color{red}Anvay:\ }#1)}}}}
\newcommand{\Filipov}[1]{{\normalsize{\textbf{({\color{green}Filipov:\ }#1)}}}}

\title [mode = title]{A Reduced Order Model for Emergent Mechanics in Woven Systems}

\author[1]{Anvay A. Pradhan}[]
\ead{anvay@umich.edu}

\author[1,2]{Evgueni T. Filipov}[]
\ead{filipov@umich.edu}

\author[1,3,4]{Talia Y. Moore}[]
\ead{taliaym@umich.edu}

\affiliation[1]{organization={Department of Mechanical Engineering, University of Michigan}, 
                city={Ann Arbor}, 
                state={Michigan}, 
                postcode={48109},
                country={USA}
                }
                
\affiliation[2]{organization={Department of Civil and Environmental Engineering, University of Michigan}, 
                city={Ann Arbor}, 
                state={Michigan}, 
                postcode={48109},
                country={USA}
                }

\affiliation[3]{organization={Department of Robotics, University of Michigan}, 
                city={Ann Arbor}, 
                state={Michigan}, 
                postcode={48109},
                country={USA}
                }

\affiliation[4]{organization={Department of Ecology and Evolutionary Biology, Museum of Zoology, University of Michigan}, 
                city={Ann Arbor}, 
                state={Michigan}, 
                postcode={48109},
                country={USA}
                }

\begin{abstract}
Woven structures exhibit rich mechanical behaviors including anisotropic stiffness, shear-induced locking, and crimp interchange that emerge purely from the geometric arrangement of individual weavers rather than from constituent material properties. 
Existing models either homogenize these interactions or resolve them at prohibitive computational cost. 
We introduce a reduced-order model that bridges this gap by representing individual weaver interactions through a system of nodes and four physically interpretable stiffness elements capturing axial deformation, in-plane uncrimping, inter-weaver shear, and frictional slip. 
Eigenvalue analysis of the unit cell confirms that the lowest-energy deformation modes correspond directly to known weave-specific phenomena, and that each element is necessary for a complete kinematic and mechanistic description. 
Element stiffness parameters are calibrated against empirical three-point bending and shear data, achieving agreement within 5\% across varied weaver widths and spacings. 
The validated model is then applied to demonstrate capabilities beyond the reach of continuum approaches including: the emergent Poisson's response arising from crimp interchange, stepwise force reduction during progressive weaver pullout, stress localization under three distinct tearing configurations, and programmable mechanical anisotropy through spatially graded weaver stiffness. 
The physical transparency and computational efficiency of the framework position it as a practical tool for the analysis and design of woven architected materials with programmable mechanical response.
\end{abstract}


\maketitle

\section{Introduction}
\label{sec:introduction}

Metamaterials and architected materials harness the geometric arrangement of individual constituent components rather than their material properties to achieve a desired mechanical performance \citep{mendhe2011metamaterial,Wang2021,XiaogangChen2011,Parsons2013,Cirio2014}. 
This geometry-driven design paradigm is powerful because it enables structures with behaviors independent of material and scale \citep{mendhe2011metamaterial,lee2024data,coulais2018characteristic}. 
Such metamaterials have been developed across a broad range of functional regimes, including acoustic \citep{peng2014acoustic,carrara2013metamaterial,wang2022mechanical}, electromagnetic \citep{watts2012metamaterial,zhang2012magnetic,butz2013one}, thermal and fluid \citep{fan2008shaped,chen2022realizing,shen2016thermal}, and structural \citep{meza2015resilient,vangelatos2019architected,injeti2019metamaterials} applications. 
Structural metamaterials in particular are able to tailor mechanical response in ways useful for vibration damping, energy absorption, and force redirection \citep{misseroni2024origami,chai2024tailoring}.

Drawing on this design philosophy, traditional artisan crafts such as knitting \citep{singal2024programming,farrell2025programmable}, knotting \citep{moestopo2023knots,goshen2025topology}, and origami/kirigami \citep{zhao2025modular,sharma2025programmable,neville2016shape}, have recently been explored as methods for creating structural metamaterials, enabling novel actuators \citep{leanza2024active,singal2024programming}, embedded sensors \citep{Sanchez2021,cheng2025wearable}, and deployable structures \citep{beatini2022integration,redoutey2021pop}. 
Weaving, among the oldest of these crafts, has likewise proven to be a versatile medium for engineered mechanical function. 
Woven architectures have been used to create shape-morphing surfaces \citep{Buckner2020,voorwinden2025multistable,liu2021robotic,Ren2021,Baek2021}, robotic actuators and grippers \citep{yang2025weaving,kang2023grasping}, and assistive devices \citep{Sanchez2021,abtew2025intelligent}. 
Building on these applications, a growing body of work has established the weave geometry itself, including the widths, spacings, and cross-sections of individual weavers, as a design variable that can be tuned for enhanced durability, high strength-to-weight ratios, programmable anisotropy, and multi-material combinations while retaining ease of manufacturing \citep{Jabbar2014,Ayres2018,tu2025corner,Wang2021}. 
Many of these advantages arise from mechanical behaviors such as anisotropic stiffness, shear-induced locking, and crimp interchange, which emerge purely from the geometric arrangement of the weavers, independent of constituent material properties. 
We refer to these emergent, topology-driven behaviors collectively as weave-specific phenomena. 
Yet whereas origami and kirigami are now supported by efficient, physically interpretable modeling frameworks that have accelerated their use as design tools, comparable tools for weaving remain limited, leaving much of this design potential difficult to predict and therefore to exploit.

Efforts to model woven systems span a similar range of fidelity. 
Most established engineering models assume that the characteristic size of individual weavers is much smaller than the overall weave \citep{Peirce1937,Kawabata1973a,Kawabata1973b,Kawabata1973c,Kawabata1992,KUWAZURU2004}. 
This assumption permits homogenization, allowing woven materials to be represented as anisotropic continua with constitutive relations aligned along the warp and weft directions \citep{King2005,Cirio2014,Parsons2010,Parsons2013}. 
Such continuum approaches enable rapid computation and capture the bulk response well, but by construction they smooth away weave-specific phenomena \citep{Parsons2010,Parsons2013,XiaogangChen2011} and can break down under large deformations or self-contact, where geometric nonlinearities dominate. 
Intermediate to these, lumped-parameter models, including bar-and-hinge formulations for origami \citep{Schenk2011,filipov2017bar,Liu2017} and more recent triangle-based formulations extended to ribbons and woven structures \citep{zhu2025lumped}, represent structured sheets with a coarse network of analytically derived elements, retaining geometric nonlinearity at far lower cost than full finite element analysis. 
To date, however, such models have largely targeted folding and contact kinematics, with element stiffnesses derived directly from constituent geometry rather than calibrated to capture the weaver-scale in-plane and crimp interactions central to weave-specific behavior. 
At the opposite extreme, high-fidelity computational methods, such as finite element analysis, resolve these behaviors accurately, including yarn-scale contact and slip, but at the cost of substantial computational resources and long runtimes \citep{Simon2024,Durville2010,Iwata2019}.

To overcome these limitations, we introduce a reduced-order model (ROM) that captures weave-specific phenomena at the level of individual weaver interactions while remaining computationally efficient. 
The model occupies the same intermediate regime as these lumped-parameter formulations, explicitly representing the geometric interactions between individual weavers within a simplified, physically interpretable framework \citep{Schenk2011,filipov2017bar,Liu2017}. 
Each weaver acts as a slender member carrying load primarily along its centerline, so its axial stretch is well-represented by a bar element; the in-plane shear and crimp couplings that a continuum panel element would otherwise have to resolve internally arise between weavers in a woven sheet and are instead represented by discrete shear and crimp springs. 
Unlike continuum approaches, the ROM preserves weave-specific behaviors by construction, and unlike high-fidelity finite element models, it enables rapid simulation and parametric exploration. 
Because element stiffnesses are treated as effective parameters rather than quantities derived from constituent geometry and material, the model must be calibrated against empirical data before it can make quantitative predictions. 
We treat this calibration as an explicit step and assess how well a single calibrated parameter set generalizes across loading conditions and weave geometries. 
This makes the framework particularly well-suited for design space exploration and parametric study of woven architected materials.
\FloatBarrier
\section{Model Formulation}
\label{sec:modelFormulation}
The ROM adopted here belongs to the family of bar-and-hinge, or node-and-spring, structural models \citep{Schenk2011,filipov2017bar,Liu2017}. 
In this class of model, a structured thin sheet is discretized into a network of axial bar members that resist stretching and rotational springs that resist bending and folding, with each node carrying only three translational degrees of freedom (DOF) and no rotational DOF \citep{Schenk2011,filipov2017bar,patil2026threenode}. 
Bending, folding, and shear kinematics are encoded through the element connectivity and the angular measures of the spring elements rather than through nodal rotations, so the full kinematic response is recovered from translations alone. 
This minimal-DOF construction is what gives the approach its efficiency: by avoiding rotational DOF and the fine meshes required by continuum elements, bar-and-hinge models capture nonlinear mechanics \citep{Liu2017,filipov2017bar}, dynamics, and coupled multi-physics behavior \citep{zhu2021multiphysics,zhuSchenkFilipov2022} at a fraction of the computational cost of full finite-element analysis. 
We adapt this framework to woven architectures by defining a unit cell and a set of element types tailored to the dominant weaver interactions, as developed in the remainder of this section.

\subsection{Geometric Formulation}

\begin{figure}[!t]
    \centering
    \includegraphics[width=0.48\columnwidth]{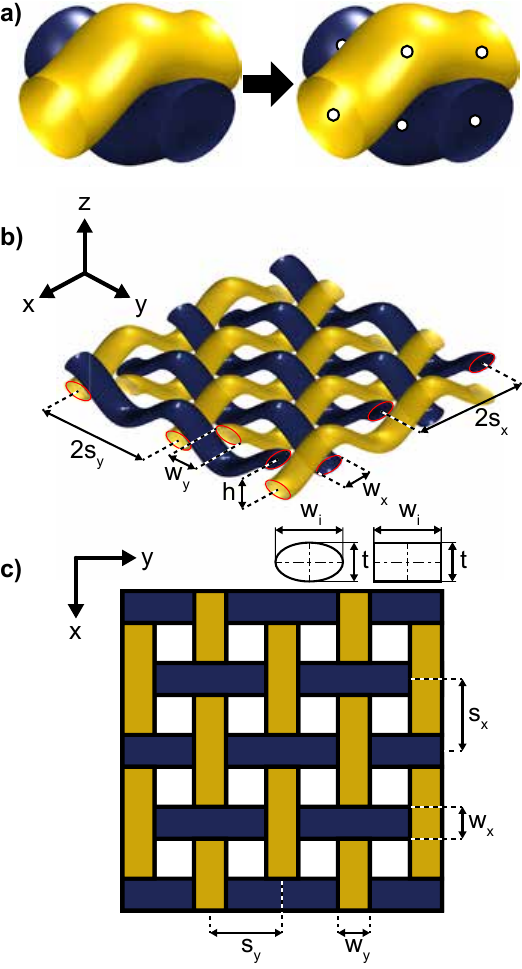}
    \caption{A plain woven sheet is modeled as a tessellation of repeating over--under unit cells, each parameterized by a compact set of geometric variables. 
    a) The over--under interlacing of two weavers, with the nodes used to define the unit cell marked. 
    Each unit cell is initially defined by six nodes aligned with the weaver centerlines; connectivity to neighboring cells reduces the description to two independent nodes per cell. 
    b) The weaver geometry is defined by the number of weavers $m \times n$, widths $w_x$ and $w_y$, spacings $s_x$ and $s_y$ (with $2s_x$, $2s_y$ the unit-cell repeat distances), and the vertical center-to-center height $h$. 
    c) Top-down view of the same parameters. 
    For visualization, weavers are rendered with elliptical or rectangular cross-sections parameterized by major axis $w_i$ and minor axis $t$: flat, ribbon-like weavers correspond to $t = 0$ and circular cross-sections to $t = w_i$. 
    The center-to-center height is the sum of the two half-thicknesses, $h = t_{\text{bottom}}/2 + t_{\text{top}}/2$, where $t_{\text{bottom}}$ and $t_{\text{top}}$ are the thicknesses of the lower and upper interlaced weavers.}
    \label{fig:unitCell}
\end{figure}

In our simulation, the weave is modeled as a tessellation of repeating over–under unit cells. 
Each unit cell contains two nodes positioned along the centerlines of the intersecting warp ($x$) and weft ($y$) weavers, as illustrated in Fig.~\ref{fig:unitCell}. 
This centerline discretization assumes the dominant mechanical response of the weave is governed by centroidal kinematics.
The precise cross-sectional shape plays a secondary role and is therefore not explicitly represented by the nodal positions.

Although cross-sectional shape is not resolved geometrically, the cross-sectional area $A$ is retained as it directly enters the bar element stiffness definition (Eq.~\ref{eq:barEnergy}).
For elliptical cross-sections, the area is given by $A = \frac{\pi}{4} w_i t$, while for rectangular cross-sections it is $A = w_i t$, where $w_i$ is the major axis width and $t$ is the minor axis thickness.

The unit cell connectivity is exploited to minimize the total number of DOF. 
Since each node is shared between neighboring cells, enforcing compatibility at the cell interfaces reduces the independent nodal description from six nodes to two nodes per unit cell. 
This reduction preserves the essential kinematics and mechanics of the weave while keeping the model compact and computationally efficient.

\subsection{Mathematical Formulation}

\begin{figure}[!t]
    \centering
    \includegraphics[width=0.99\columnwidth]{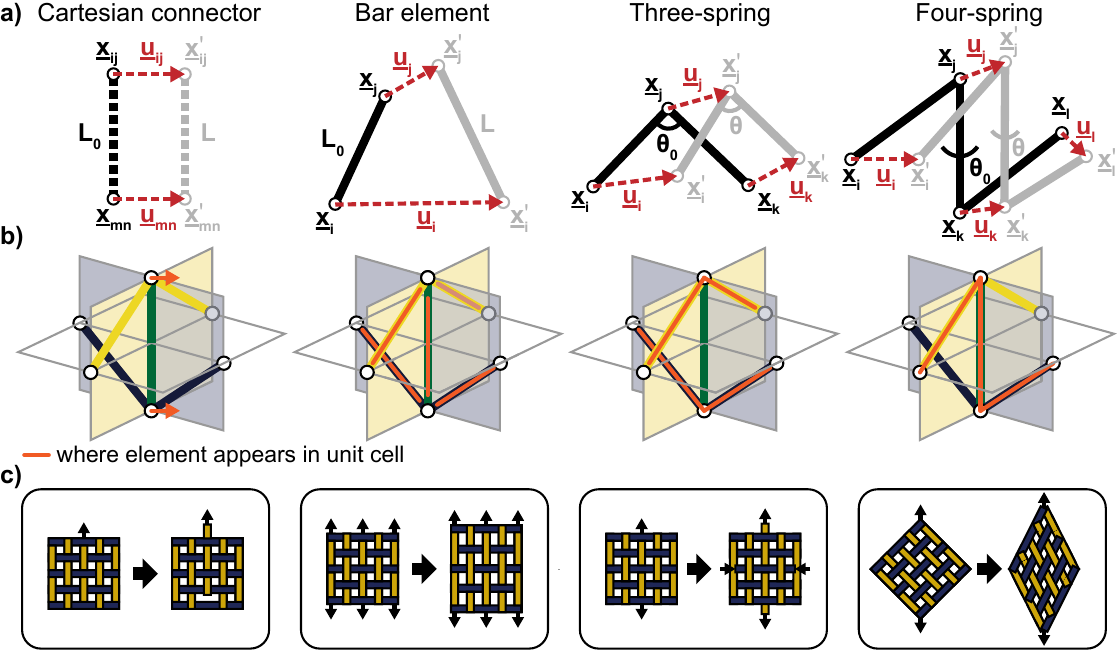}
    \caption{The reduced-order model discretizes each unit cell into four stiffness elements that together capture the principal weave-specific mechanical behaviors: (1) a Cartesian connector capturing pseudo-frictional slip, (2) a bar element capturing axial stretch, (3) a three-node spring capturing weaver crimping, and (4) a four-node spring capturing inter-weaver shear. 
    a) Kinematic definition of each element: reference node positions $\underline{x}$, deformed positions $\underline{x}' = \underline{x} + \underline{u}$, and the length measure ($L_0$, $L$) or angle measure ($\theta_0$, $\theta$) whose deviation from its reference value each element penalizes. 
    b) Where each element appears within the unit cell (highlighted). For clarity only one four-node spring is shown, though four are defined per unit cell. 
    c) The macroscopic weave deformation mode governed by each element, shown as a reference-to-deformed pair: slip, axial stretch, crimping, and shear, respectively.}
    \label{fig:stiffnessElems}
\end{figure}

Inter-weaver interactions within and between cells are represented through four physically interpretable stiffness elements: (1) a Cartesian connector capturing pseudo-frictional slip, (2) a bar element capturing axial stretch, (3) a three-node spring capturing weaver crimping, and (4) a four-node spring capturing inter-weaver shear.

Each node $i$ has reference position $\underline{x}_i = \{x_{i1}, x_{i2}, x_{i3}\}$ and a displacement vector $\underline{u}_i = \{u_{i1}, u_{i2}, u_{i3}\}$ such that its deformed position is $\underline{x}_i' = \underline{x}_i + \underline{u}_i$.
The stored energy of each element ($\Pi$) is written as a quadratic penalty on the deviation of a single geometric measure from its reference value:
\begin{gather} 
    \Pi_{cart} = \tfrac{1}{2}\, k_{cart} \left( L_{cart}-L_{0,cart} \right)^2, \\
    \Pi_{bar} = \tfrac{1}{2}\, \frac{E_{bar}A_{bar}}{L_{0,bar}} \left( L_{bar}-L_{0,bar} \right)^2 \label{eq:barEnergy}, \\
    \Pi_{3spr} = \tfrac{1}{2}\, k_{3spr} \left( \cos\theta_{3spr}-\cos\theta_{0,3spr} \right)^2, \\
    \Pi_{4spr} = \tfrac{1}{2}\, k_{4spr} \left( \cos\theta_{4spr}-\cos\theta_{0,4spr} \right)^2.
\end{gather}
In each of these energy expressions, the reference and deformed values of the length measure are denoted by a subscripted $L_{0,(\cdot)}$ and $L_{(\cdot)}$, while the reference and deformed values of the angle measure are denoted by a subscripted $\theta_{0,(\cdot)}$ and $\theta_{(\cdot)}$. All length and angle measures are evaluated directly from the deformed nodal positions $\underline{x}_i'$, which makes every element energy a function of the nodal DOF. 
For the Cartesian connector, $L_{cart}$ is the separation between the two coupled nodes along the connector direction, $L_{0,cart}$ is its reference value, and $k_{cart}$ is an effective spring stiffness. 
The connector is formulated as a direct linear spring, rather than an axial $E A / L_{0}$ member, because it connect nodes that share an in-plane location (the stacked warp and weft layers), for which the reference separation $L_{0,cart}$ along the coupled direction vanishes and an $E A / L_{0,cart}$ form would be singular. 
For the bar element, $L_{bar}$ is the Euclidean distance between its two end nodes and $L_{0,bar}$ its undeformed length, $E_{bar}$ is the weaver elastic modulus, and $A_{bar}$ is the cross-sectional area introduced in the geometric formulation above. 
For the three-node spring, $\theta_{3spr}$ is the opening angle of a single weaver given by a central node and its two adjacent segments, with reference value $\theta_{0,3spr}$ and rotational stiffness $k_{3spr}$.
Our three-node spring element is the three-dimensional generalization of the planar two-bar torsional spring of \citet{patil2026threenode}, with the in-plane perp-dot construction replaced by a cross-product form valid for arbitrary spatial configurations. 
For the four-node spring, $\theta_{4spr}$ is the dihedral angle between the planes of two crossing weavers about their shared edge, with reference value $\theta_{0,4spr}$ and rotational stiffness $k_{4spr}$.
The spring energies are expressed through $\cos\theta_{(\cdot)}$ rather than $\theta_{(\cdot)}$ directly so that the energy and its derivatives remain well-defined at flat configurations ($\theta_{(\cdot)} = 0$ or $\theta_{(\cdot)} = \pi$), where the gradient of the angle is singular which allows an initially flat, uncrimped sheet to be assembled and solved.
The stiffness parameters $k_{cart}$, $E_{bar}$, $k_{3spr}$, and $k_{4spr}$ each characterize the resistance of the corresponding element to its associated deformation mode. 
Rather than being derived from the detailed weaver geometry, they are treated as effective constitutive properties of the weave and are calibrated using experimental measurements (Section~\ref{sec:stiffnessCalibration}).

For each element, the internal force vector ($f_{local}$) and local stiffness matrix ($K_{local}$) follow as the first and second derivatives of its stored energy with respect to the nodal displacements ($u_{ij}$ \& $u_{mn}$):
\begin{equation}
    f_{local} = \frac{\partial \Pi}{\partial u_{ij}},
    \qquad
    K_{local} = \frac{\partial^2 \Pi}{\partial u_{ij}\,\partial u_{mn}} .
    \label{eq:forceStiffness}
\end{equation}
The full analytical expressions for these derivatives, together with the finite central-difference approximation used to verify them, are derived in Appendix A.

The local matrices are assembled into the global stiffness matrix $K_{global}$ by a standard direct-stiffness procedure: each node is assigned a global DOF index, and the entries of $K_{local}$ are summed into the corresponding rows and columns of $K_{global}$. 
Because nodes are shared between neighboring unit cells, this assembly automatically enforces displacement compatibility across cell interfaces. 
Displacement constraints are limited to fixed (homogeneous Dirichlet) supports; no periodic or multi-point constraints are imposed. 
The system is partitioned into free and fixed DOF, the fixed DOF are removed by row-and-column elimination, and applied loads are placed in the corresponding free-DOF entries of the global force vector $F_{global}$. 
For a linear analysis, the free-DOF displacements $\Delta_{global}$ satisfy the linear system
\noindent
\begin{equation}
    K_{global}\, \Delta_{global} = F_{global},
    \label{eq:linearSolve}
\end{equation}
which is solved by direct factorization of $K_{global}$ rather than by explicitly forming its inverse. 
The assembled global system is solved within an incremental-iterative framework that accommodates large geometric nonlinearities.
The implementation provides forward-Euler load control as a lightweight predictor, Newton--Raphson iteration for quadratic convergence under monotonic loading, and arc-length continuation for tracing equilibrium paths.
\FloatBarrier
\section{Element Validation}
\label{sec:elementValidation}

To capture the fundamental mechanics of woven structures, each stiffness element must be validated independently, as each represents a distinct physical behavior within the weave.
We first confirm the correct implementation of each element by comparing its load response against an analytical solution.
We then combine the validated elements into a single unit cell and perform an eigenvalue analysis to identify the elastic modes of the system, revealing how the combined elements reproduce the expected deformation modes of woven systems.

\subsection{Individual Element Characterization}

\subsubsection{Cartesian Connector}
Cartesian connectors introduce stiffness between specific DOFs of adjacent nodes. 
For example, the x-direction displacement of one node can be coupled to the x-direction displacement of another, allowing control over relative slip between weavers. 
To evaluate its stiffness response, two nodes were defined with a single Cartesian connector linking their corresponding DOF. 
One node was fixed in all three translational directions and the other was allowed to move freely. 
A unit load was applied sequentially in each Cartesian direction and the displacement recorded. 
This analysis was repeated for each possible DOF coupling configuration. 
In every case, the connector exhibited linear stiffness in the coupled direction and zero stiffness in the orthogonal directions, confirming that the Cartesian connector accurately captures the intended slip behavior.

\subsubsection{Bar Element}
The axial bar is a standard element of bar-and-hinge and truss formulations, with a well-established stiffness \citep{Schenk2011,filipov2017bar,mcguire2000}. 
Unlike the Cartesian connector, the bar introduces stiffness along all three translational DOF simultaneously, representing axial stiffness along the length of the element.
Defining a bar element is equivalent to defining three Cartesian connectors, one per translational direction.
As an implementation check, a single bar was loaded between a fixed and a free node in each coordinate direction successively; the response was linear in all three, confirming the element was assembled and integrated correctly.

\subsubsection{Three-Node Spring}
Three-node springs introduce rotational stiffness about the angle subtended by the central node, enabling the model to capture crimp interchange as weavers are pulled in tension. 
This element follows the three-node torsional spring of \citet{patil2026threenode}, extended here to three dimensions.
To validate this element, two of the three nodes were fixed in all three translational DOF.
The third was constrained in the x-direction to prevent out-of-plane motion, leaving it free to move in the $y$-$z$ plane.
The free node was positioned at discrete locations along the expected arc and subjected to a vertical load in the z-direction.
The normalized displacement response follows a single cosine period over the range $\theta_0 \in [0, \pi]$, reaching maximal displacement at $\theta_0 = 0$ and $\theta_0 = \pi$ and vanishing at $\theta_0 = \pi/2$.
This result was consistent with the analytical solution for a rotational spring and confirms that the three-node spring accurately captures crimp interchange behavior.

\subsubsection{Four-Node Spring}
Four-node springs introduce rotational stiffness between two planes, each defined by three nodes, where the two shared nodes form the rotation axis between the planes. 
This definition enables the model to capture inter-weaver shear between intersecting warp and weft weavers. 
The validation procedure was identical to that of the three-node spring, with three nodes fixed and the fourth constrained in the x-direction, subjected to a vertical load in the z-direction at discrete locations along the expected arc.
The normalized displacement response exhibits the same single cosine period over $\theta_0 \in [0, \pi]$, confirming that the four-node spring accurately captures inter-weaver shear behavior.

\subsection{Unit Cell Elastic Modes}
With all elements validated individually, a single unit cell was constructed from six nodes, yielding 18 total DOF. 
Rigid-body motions were suppressed by imposing boundary conditions on the unit cell, reducing the system to its independent elastic modes. 
A minimal six-constraint support set is sufficient to remove rigid-body motion (three translational and three rotational) but treats the in-plane $x$ and $y$ directions unequally because the constraints that suppress rotation are not symmetric between the two axes. 
This asymmetry artificially splits the energies of the two in-plane slip modes (modes 3 and 4), which the symmetry of the weave requires to be degenerate. 
A seventh constraint is therefore imposed to restore a symmetric boundary treatment, recovering the expected degeneracy of the slip modes.
The resulting distribution of eigenvalues reflects the intrinsic stiffness of the weave rather than an artifact of the support scheme.
The global stiffness matrix of this constrained unit cell was then subjected to eigenvalue analysis to identify the modes and their associated energy costs.

\begin{figure}[!t]
    \centering
    \includegraphics[width=0.99\columnwidth]{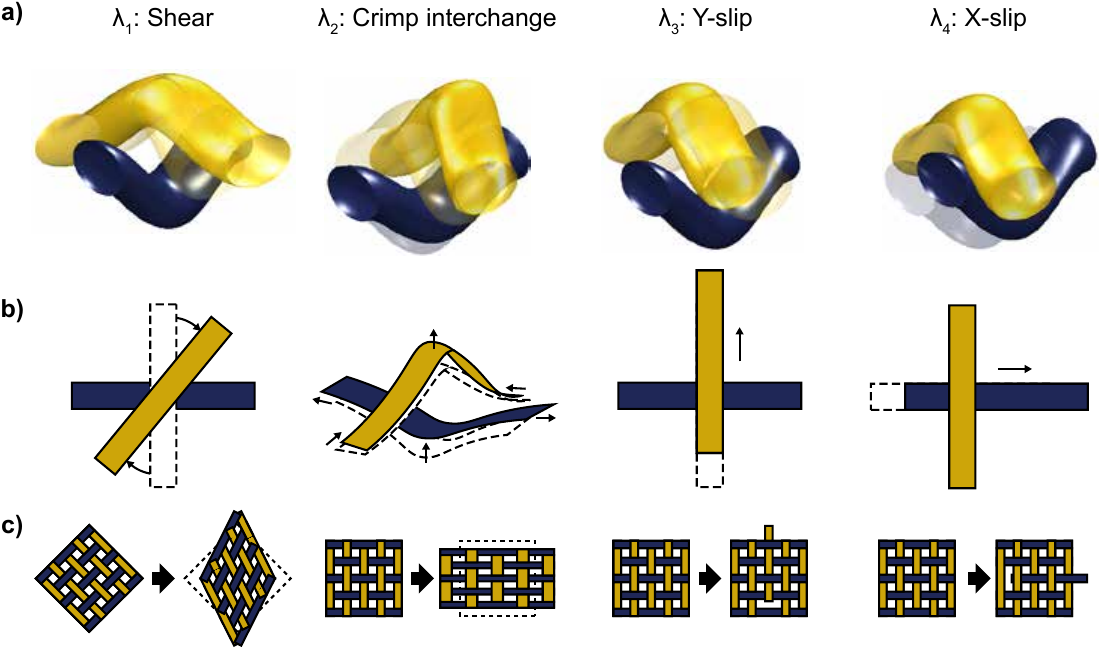}
    \caption{Eigenvalue analysis of the unit cell shows that its lowest-energy elastic modes correspond to recognizable weave-specific phenomena, supporting the necessity of the full element set. 
    a) The four lowest-energy elastic modes of the constrained unit cell, ordered by increasing eigenvalue, with the undeformed configuration shown as a shadow. The modes correspond to shear, crimp interchange, $y$-slip, and $x$-slip; the two slip modes form a degenerate pair related by the in-plane symmetry of the weave. 
    b) Simplified sketches of the individual weaver behavior associated with each mode. 
    c) The corresponding global deformation of the weave produced by each mode. 
    These four modes are the lowest of the eleven independent elastic modes of the unit cell; the appearance of distinct shear, crimp, and slip modes confirms that the corresponding stiffness elements are each required to represent the full elastic response.}
    \label{fig:eigenmodes}
\end{figure}

The eigenvalue analysis revealed two key observations. First, there is a two order of magnitude difference in energy between modes 4 and 5, as seen in Table \ref{tables:eigenmodes}, indicating that the first four modes are energetically preferred and therefore most likely to occur under loading. 
Second, plotting the modes (Fig.~\ref{fig:eigenmodes}) reveals that the four lowest-energy deformations correspond directly to the weave-specific phenomena of shear, crimp interchange, and slip, while the higher modes represent coupled combinations of these fundamental behaviors and more complex stretching behaviors. 
Together, these observations confirm that the reduced-order model naturally prioritizes the physically meaningful deformation modes of woven structures.

\begin{table}[]
\caption{Eigenvalues of Unit Cell}
\label{tables:eigenmodes}
\centering
\begin{tabular}{|l|l|l|}
\hline
\textbf{Mode $i$} & \textbf{Eigenvalue $\lambda_i$} & \textbf{Deformation}      \\ \hline
1               & 50                                & Shear                     \\ \hline
2               & 314                               & Crimp interchange         \\ \hline
3               & 841                               & Y-slip                      \\ \hline
4               & 841                               & X-slip                      \\ \hline
5               & 12587                             & Compression               \\ \hline
6               & 66811                             & Crimp interchange \& Slip \\ \hline
7               & 69439                             & Crimp interchange \& Slip \\ \hline
8               & 102269                            & Crimp interchange \& Slip \\ \hline
9               & 184628                            & Stretching                \\ \hline
10              & 2037030                           & Stretching                \\ \hline
11              & 4018271                           & Stretching                \\ \hline
\end{tabular}
\end{table}

The unit cell analysis was also used to confirm that each stiffness element is necessary for a complete representation of the weave mechanics. 
Each element was independently removed and the resulting eigenvalues inspected. 
In every case, removal of a single element introduced one or more zero-energy modes, indicating that an otherwise-constrained deformation was no longer being captured and confirming that no element is redundant. 
Furthermore, the deformation of the newly introduced zero-energy modes identified the physical behavior governed by each element and was consistent with the individual element validations above. 
This one-to-one correspondence between element type and deformation mode provides a physically grounded basis for calibrating the stiffness parameters of each element independently to match observed weave behavior.
\FloatBarrier
\section{Stiffness Calibration}
\label{sec:stiffnessCalibration}

While the element validation confirmed correct implementation and necessity of each stiffness element, the predictive capability of the model depends on assigning physically meaningful stiffness values. 
For bar elements, the modulus of elasticity, $E_{bar}$, and cross-sectional area, $A_{bar}$, can be measured directly from the weaver material. 
However, the stiffness parameters of the Cartesian connector, three-node spring, and four-node spring cannot be measured on a single weaver in isolation.
Therefore, empirical testing was conducted on woven samples under common loading scenarios, and a data-driven optimization scheme was developed to determine these parameter values.

\subsection{Empirical Testing}
To isolate the geometry-driven mechanical response of the weave from confounding material effects, samples must be fabricated from a linearly elastic and isotropic material. 
Natural fibers, such as cotton or wood ribbons, were therefore not suitable, as their anisotropic and nonlinear material behavior would obscure the geometric contribution to the overall response.
\\
\subsubsection{Sample Preparation}
\begin{figure}[!t]
    \centering
    \includegraphics[width=0.48\columnwidth]{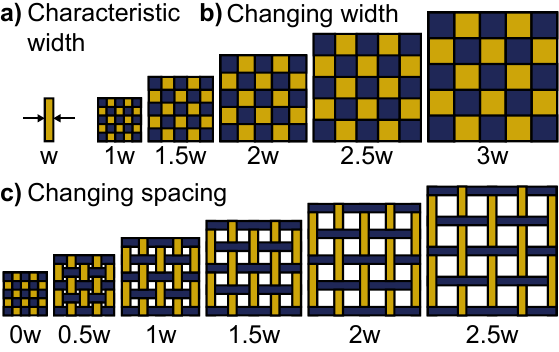}
    \caption{Physical test samples were designed using a normalized parametrization that varies weaver width and spacing independently and expresses both in units of a common characteristic width, enabling direct comparison with simulation predictions. 
    a) A characteristic width $w$ is defined, and all sample dimensions are expressed as multiples of $w$, generalizing the sample set across scales and facilitating comparison with computational predictions. 
    b) Weaver width was varied independently across samples, from $1w$ to $3w$, at fixed spacing. 
    c) Weaver spacing was varied independently across samples, from $0w$ (adjacent weavers in contact) to $2.5w$, at fixed width. 
    Both sweeps are expressed as multiples of the characteristic width $w$.}
    \label{fig:samples}
\end{figure}

Samples were fabricated using 7.5~mil thick Mylar (Milky translucent PET
template, Frienda, Shenzhen, China). 
Strips were cut using a machine blade (Silhouette Cameo, Lindon, UT, USA) rather than a laser cutter, to avoid heat-affected zones that would alter the material properties along the cut edge.
Each strip had a characteristic width of 1~cm and the overall weave pattern was fixed at a size of 5$\times$5 weavers.
Samples were then fastened together using 8~mm head diameter brass paper brads (Bremorou, Casper, WY, USA). 
The samples were designed to vary weaver width and weaver spacing independently, as shown in Fig.~\ref{fig:samples}.
For varying width, samples were of size 5$\times$5, 7.5$\times$7.5, 10$\times$10, 12.5$\times$12.5, and 15$\times$15~cm, while for varying spacing, samples were of size 5$\times$5, 7$\times$7, 9$\times$9, 11$\times$11, 13$\times$13, and 15$\times$15~cm.

\subsubsection{Testing Methodology}
\begin{figure}[!t]
    \centering
    \includegraphics[width=0.48\columnwidth]{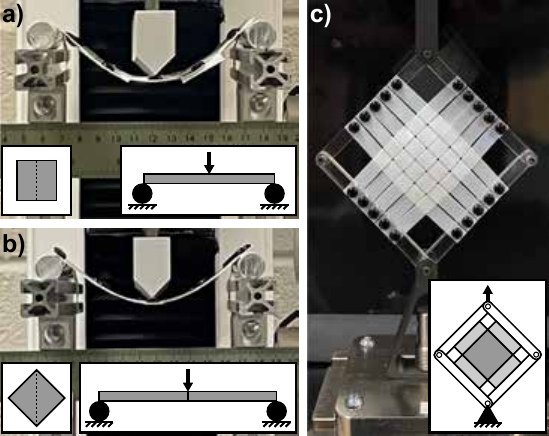}
    \caption{Three empirical loading configurations were used to characterize the bending and shear response of the woven samples. 
    Testing setups for a) three-point bending in the perpendicular orientation (load aligned with the weaver axes), b) three-point bending in the bias orientation (load at $45^\circ$ to the weaver axes), and c) picture-frame shear testing. 
    Insets in (a) and (b) indicate the sample orientation relative to the loading direction.}
    \label{fig:tests}
\end{figure}

Three loading configurations were used to characterize the mechanical response of the woven samples (Fig.~\ref{fig:tests}): 
(1) three-point bending along the centerline, referred to as the perpendicular orientation (Fig.~\ref{fig:tests}a), 
(2) three-point bending along the diagonal axis, referred to as the bias orientation (Fig.~\ref{fig:tests}b), and 
(3) shear loading (Fig.~\ref{fig:tests}c). 
Together, these configurations subjected the samples to bending, shear, and friction: the three behaviors the model is designed to capture. 
Simple uniaxial tension and compression along the weaver directions were excluded, as the material response would dominate the geometry-driven mechanical response under such loading. 
Torsional loading was similarly excluded, as it is outside the scope of the current model. 
Each loading scenario also corresponds to a closed-form analytical solution for a continuous, non-woven sample, facilitating direct comparison with simulation predictions.

Separate samples were fabricated for three-point bending and shear testing. 
For both three-point bending configurations, a 10~N load cell attached to a linear load frame (M5-10 digital force gauge, Mark-10, Copiague, NY, USA) applied a displacement-controlled load through a custom 3D-printed fixture designed to distribute the load evenly along a single line. 
For the perpendicular orientation, samples were supported on two rollers with the sample edges aligned with the roller axes. 
For the bias orientation, the sample corners were positioned at a distance of 10\% of the mid-span from the rollers. 
Each sample was tested three times per configuration.
As no plastic deformation was observed between consecutive trials, samples were reused across repetitions. 

For shear loading, the same load cell and load frame were used. 
A custom acrylic frame was laser cut and assembled to constrain the sample corners and enforce pure shear kinematics consistently across all samples. 
Woven samples were then fastened to the frame using the brass paper brads. 
Prior to sample testing, the frictional contributions of the frame alone and of a single brad were quantified independently and subtracted from all subsequent measurements to isolate the mechanical response of the weave. 
The weave was first cyclically sheared three times, with the upper and lower strain bounds determined by the onset of out-of-plane buckling due to shear locking. 
The sample was then cycled three additional times within $\pm20\%$ strain of the starting orientation.

\subsubsection{Data Normalization}
Because sample dimensions varied across tests, the raw load--displacement data were mapped onto normalized response measures so that results from differently sized samples could be compared directly. 
These measures are deliberately stress-, strain-, and modulus-like in form and units, but they are \emph{normalizations} chosen to remove sample-size dependence, not constitutive properties of the weave.
Their validity is established empirically rather than derived from continuum theory. 
We denote them with hatted symbols to make this distinction explicit. 

For three-point bending, the normalized bending stress $\hat\sigma_{\mathrm{b}}$ and normalized bending strain $\hat\epsilon_{\mathrm{b}}$ are defined as
\begin{equation}
    \hat\sigma_{\mathrm{b}} = \frac{P\,L_{\mathrm{eff}}\,t}{8 I_z},
    \qquad
    \hat\epsilon_{\mathrm{b}} = \frac{d}{L_{\mathrm{eff}}},
\end{equation}
and the normalized bending modulus
\begin{equation}
    \hat E_{\mathrm{b}} = \frac{\hat\sigma_{\mathrm{b}}}{\hat\epsilon_{\mathrm{b}}}
    = \frac{P\,L_{\mathrm{eff}}^{2}\,t}{8 I_z\,d}.
\end{equation}
Here $P$ is the applied load, $d$ the midspan displacement, $t$ the sample thickness, and $I_z$ the second moment of area of the nominal rectangular cross-section. 
The effective span is $L_{\mathrm{eff}} = L$ for the perpendicular orientation and $L_{\mathrm{eff}} = \sqrt{2}\,L$ for the bias orientation, the $\sqrt{2}$ reflecting the longer diagonal load path, where $L$ is the roller-to-roller support span. 
This form follows the three-point bending configuration of ASTM~D790 \citep{astmD790}, from which it is adapted. 

For shear, the normalized shear stress $\hat\sigma_{\mathrm{s}}$ and normalized shear strain $\hat\epsilon_{\mathrm{s}}$ are defined as
\begin{equation}
    \hat\sigma_{\mathrm{s}} = \frac{P}{A_{s}},
    \qquad
    \hat\epsilon_{\mathrm{s}} = \frac{d}{L_{\mathrm{eff}}},
\end{equation}
with normalized shear modulus
\begin{equation}
    \hat E_{\mathrm{s}} = \frac{\hat\sigma_{\mathrm{s}}}{\hat\epsilon_{\mathrm{s}}}
    = \frac{P\,L_{\mathrm{eff}}}{A_{s}\,d}.
\end{equation}
Here $d$ is the diagonal displacement and $L_{\mathrm{eff}}$ is the initial diagonal distance of the square shear sample. 
The term $A_{s} = L_{\mathrm{eff}}\,t$ is the effective cross-sectional area through which the diagonal load is transferred, taken as the transverse diagonal length times the sample thickness $t$. 
The shear loading follows the picture-frame method of ASTM~D8067 \citep{astmD8067}, in which a square sample is loaded along its diagonal so that the fixture deforms from a square into a paralleogram.
This approach produces a shear-dominated deformation with $A_{s}$ defined as above and $\hat\sigma_{\mathrm{s}}$ consistent with the apparent shear stress of that standard. 

To confirm that these definitions remove sample-size dependence, five continuous (non-woven) samples were fabricated from the same Mylar stock used for the weaves and were tested in each loading configuration at several sizes.
Applying the normalizations collapsed the load--displacement responses of all sizes onto a single curve in each configuration, confirming that the normalized measures are size-independent for a known continuous material. 
Having established this, the same normalizations were applied to the woven samples for comparison across geometries and loading conditions.

\subsection{Data-driven Parameter Calibration}
For each empirical stress--strain curve, a linear fit over the elastic regime was used to extract the corresponding normalized stiffness ($\hat E_{\mathrm{b}}$ for bending, $\hat E_{\mathrm{s}}$ for shear). 
These stiffnesses were plotted against the varied geometric parameter (weaver width or spacing) to establish a scaling trend for each loading configuration. 
Because the normalizations were constructed to remove sample-size dependence for a continuous material, the normalized stiffness of a continuous reference sample is geometry-independent.
Any residual variation of normalized stiffness with weave geometry therefore reflects weave-specific behavior, and it is this trend that the model is calibrated to reproduce.
The reduced-order model is conservative and quasi-static by construction: every element energy is a potential, so the assembled response is single-valued and path-independent. 
This places frictional hysteresis, a dissipative phenomenon with no associated potential, outside the scope of the present formulation, which targets the elastic, geometry-driven response of the weave rather than the frictional dissipation between weavers. 
The experimental shear response is distinctly hysteretic, tracing a closed loop under cyclic loading (Fig.~\ref{fig:shearTests}b). 
To calibrate the static model against these data while representing only what it is designed to capture, each measured shear loop is sectioned into its four monotonic regimes (the bottom, top, right, and left segments of the loop), and each segment is treated as a separate, single-valued calibration target.
The model is thus fit to the individual elastic branches of the loop, which together span the loading and unloading behavior, rather than to the dissipative loop as a whole.

This sectioning defines the set of calibration cases, $\mathcal{C}$. 
Three-point bending is non-hysteretic and contributes four cases: two orientations (perpendicular and bias), each swept over two geometric parameters (width and spacing). 
Shear contributes eight: the four loop regimes, each swept over the same two parameters. 
This yields $|\mathcal{C}| = 12$ empirical scaling trends in total. 

\begin{figure}[!t]
    \centering
    \includegraphics[width=0.48\columnwidth]{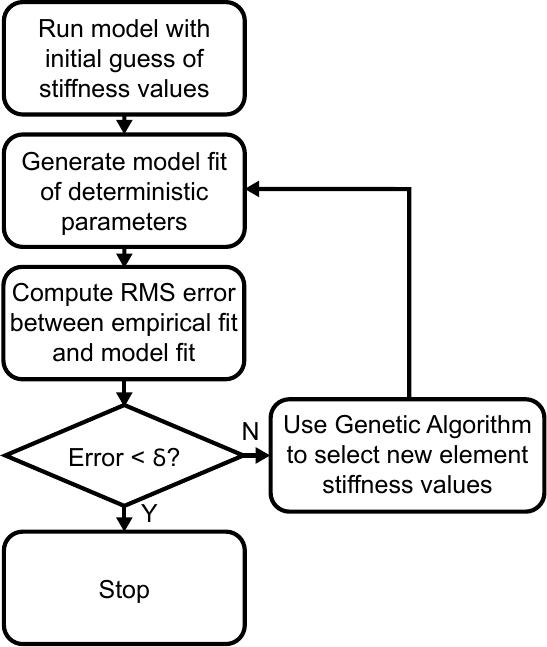}
    \caption{The parameter-calibration loop used to identify element stiffness values that best reproduce the empirical stiffness trends. 
    Starting from an initial stiffness guess, the model is run and its output is fit to obtain the simulated trend. 
    The root-mean-square (RMS) error between the empirical fit and the model fit is computed and used as the calibration metric. 
    When this error exceeds a predefined threshold $\delta$, a genetic algorithm proposes a new set of stiffness values and the loop repeats; otherwise the calibration terminates.}
    \label{fig:blockDiagram}
\end{figure}

The bar modulus, $E_{bar}$, is fixed from the known weaver material property and is not calibrated, leaving three free stiffnesses, $\mathbf{x} = \{k_{cart},\, k_{3spr},\, k_{4spr}\}$. 
For each case, the model was run under the same loading and boundary conditions as the corresponding experiment, and a scaling trend was extracted from the simulated stress--strain curves using the normalization scheme above. 
The root-mean-square (RMS) error between the simulated and empirical trends was computed and compared against a convergence threshold $\delta$ (Fig.~\ref{fig:blockDiagram}).
The coefficient of determination $R^2$ of each fit was retained as a reported goodness-of-fit diagnostic. 
If the threshold was exceeded, a genetic algorithm proposed a new stiffness set and the process repeated until convergence. 

\subsubsection{Distributed Parameter Calibration}
The optimization was first run independently for each of the 12 cases, yielding 12 specialized parameter sets, each calibrated to a single test condition:
\begin{equation}
    \mathbf{x}^{*}_{c} = \arg\min_{\mathbf{x}} 
    \bigl\| y_{\mathrm{model}}^{(c)}(\mathbf{x}) - y_{\mathrm{emp}}^{(c)} \bigr\|^{2},
    \qquad c \in \mathcal{C}.
\end{equation}
Here $y_{\mathrm{emp}}^{(c)}$ is the empirical fit for case $c$, that is, the curve fitted to the measured normalized stiffness ($\hat{E}_b$ for bending or $\hat{E}_s$ for shear) as a function of the swept geometric parameter (weaver width or spacing); $y_{\mathrm{model}}^{(c)}(\mathbf{x})$ is the corresponding trend produced by the model for a given stiffness set $\mathbf{x} = \{k_{cart}, k_{3spr}, k_{4spr}\}$; the norm $\lVert\,\cdot\,\rVert^{2}$ is the squared residual between the two trends taken over the geometric-parameter sweep; $\mathbf{x}^{*}_{c}$ is the stiffness set that minimizes this residual for case $c$; and $\mathcal{C}$ is the set of 12 calibration cases.
A convergence threshold of $\delta = 1\%$ was used. 
The resulting stiffness values are reported in Table~\ref{tables:opt}. 
These specialized fits establish the best achievable agreement for each condition in isolation and serve as a baseline against which the generalized model below is assessed. 
The per-regime shear fits also illustrate the treatment of the hysteresis loop by the model directly: the four-node spring stiffness $k_{4spr}$ differs by roughly two orders of magnitude between the loading and unloading branches (left, right) and the plateau branches (top, bottom), a spread that a single conservative stiffness cannot reconcile and that stands in for the frictional dissipation the model does not represent. These 12 individual models are used to produce the simulation results in Figures~\ref{fig:threePointBendTests} and~\ref{fig:shearTests}.

\subsubsection{Lumped Parameter Calibration}
To obtain a single generalizable model, all 12 cases were calibrated simultaneously against one shared stiffness set. 
This is a multi-objective fit: the cost function is a weighted linear combination of the squared errors between every simulated and corresponding empirical trend,
\begin{equation}
    \mathbf{x}^{*} = \arg\min_{\mathbf{x}} 
    \sum_{c \in \mathcal{C}} w_{c}\, 
    \bigl\| y_{\mathrm{model}}^{(c)}(\mathbf{x}) - y_{\mathrm{emp}}^{(c)} \bigr\|^{2},
\end{equation}
where $w_{c}$ is the scalar weight assigned to case $c$ and $\mathbf{x}^{*}$ is the single stiffness set that minimizes the total weighted residual across all 12 cases.
A single parameter set cannot match every condition as closely as a fit specialized to that condition alone, so the convergence threshold was relaxed to $\delta = 5\%$, reflecting the accuracy--generality trade-off inherent in describing all behaviors with one set of stiffnesses. 
Consistent with this, the lumped stiffnesses in Table~\ref{tables:opt} fall between the corresponding per-condition fits rather than matching any one of them. 
We retained uniform weights ($w_{c} = 1$) across all 12 cases. 
Because eight of the twelve trends derive from shear and only four from bending, the shear residuals constitute two-thirds of the total cost, and the lumped fit is consequently biased toward reproducing shear behavior at the expense of bending. 
This imbalance is a limitation of the present calibration.
A weighting that equalizes the contribution of the two loading modes (for example, $w_{c}$ inversely proportional to the number of trends per mode) would mitigate it and is a natural refinement for future work.

\subsection{Calibration Results}
\begin{figure}[!t]
    \centering
    \includegraphics[width=0.85\columnwidth]{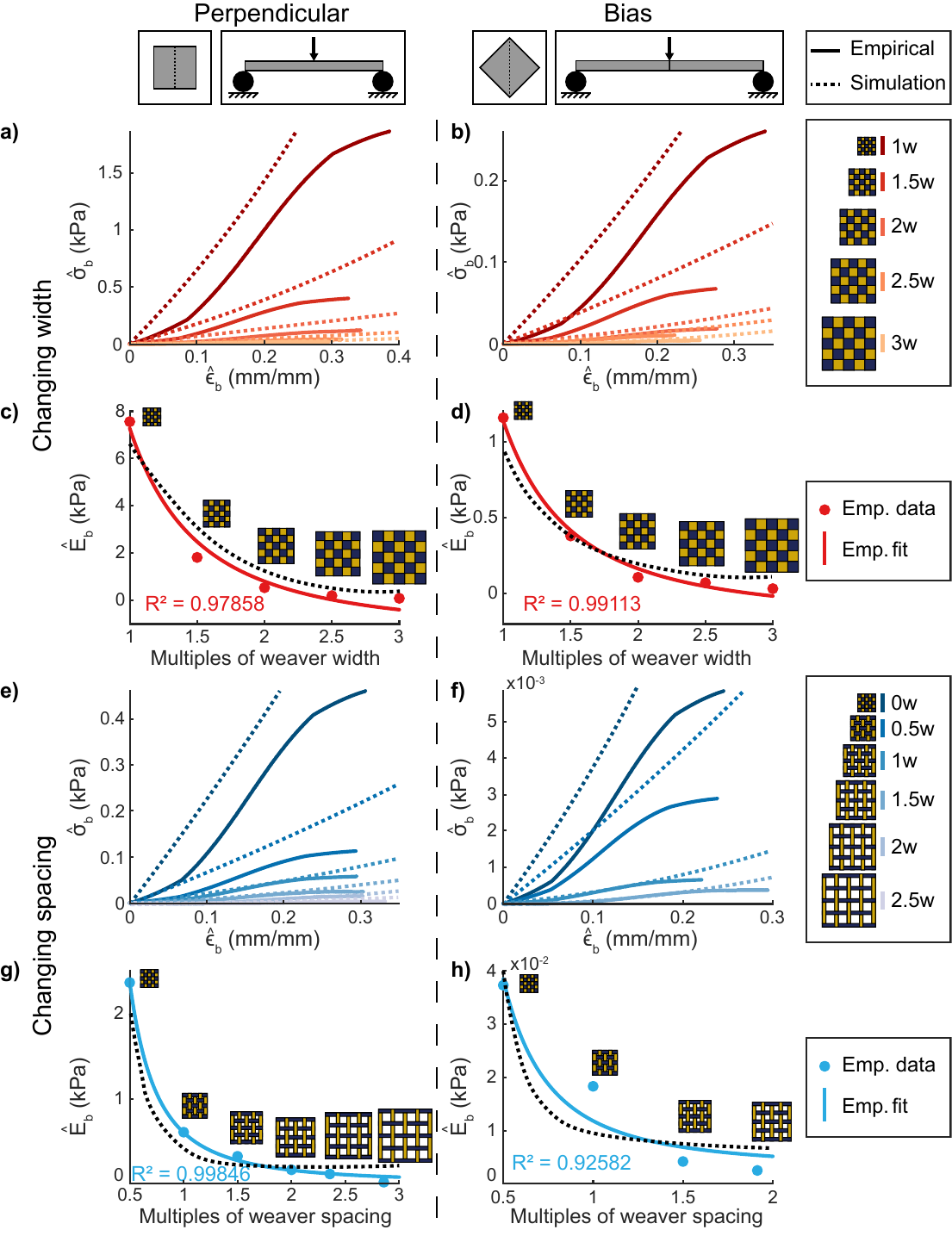}
    \caption{The distributed calibration reproduces the empirical three-point bending response across both loading orientations and both geometric sweeps to within the 1\% calibration threshold. 
    Results are shown for the perpendicular (left column) and bias (right column) orientations. 
    In the stress--strain panels, solid curves are empirical measurements and dotted curves are the calibrated simulation. 
    In the modulus panels, points are empirical data, the colored curve is the empirical fit, and the dotted black curve is the simulation; $R^2$ quantifies the agreement of the empirical fit with the data. 
    a, b) Normalized bending stress $\hat\sigma_b$ versus normalized bending strain $\hat\epsilon_b$ for varying weaver width, in the perpendicular and bias orientations. 
    c, d) Normalized bending modulus $\hat E_b$ versus weaver width, in the perpendicular and bias orientations. 
    e, f) Normalized bending stress $\hat\sigma_b$ versus normalized bending strain $\hat\epsilon_b$ for varying weaver spacing, in the perpendicular and bias orientations. 
    g, h) Normalized bending modulus $\hat E_b$ versus weaver spacing, in the perpendicular and bias orientations.}
    \label{fig:threePointBendTests}
\end{figure}

\begin{figure}[!t]
    \centering
    \includegraphics[width=0.85\columnwidth]{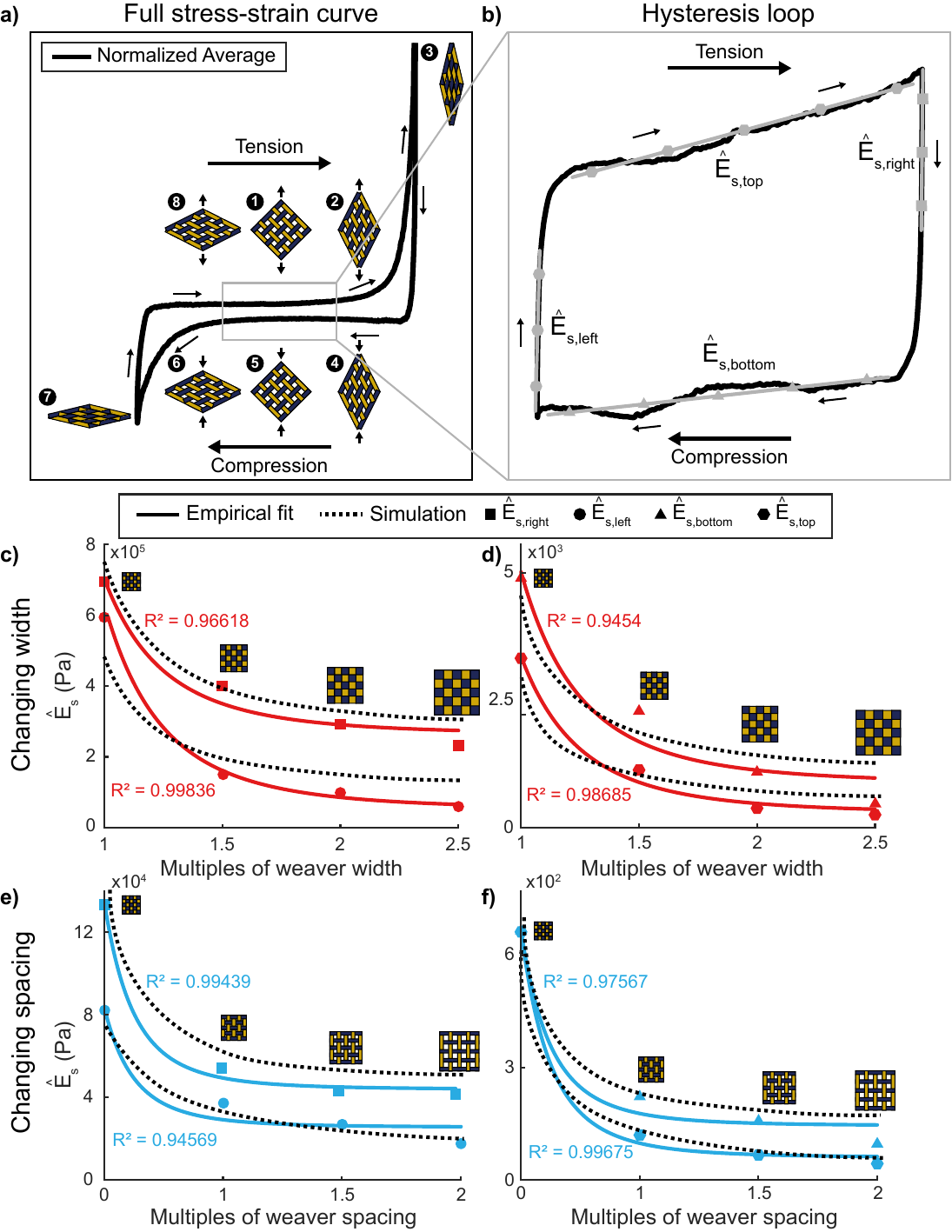}
    \caption{The distributed calibration reproduces the empirical shear stiffness trends to within the 1\% calibration threshold, demonstrating the framework's ability to accommodate piecewise-linear models representing distinct mechanical regimes. 
    Because the static model cannot reproduce a hysteresis loop, each measured loop is sectioned into four monotonic branches ($\hat E_{s,\text{right}}$, $\hat E_{s,\text{left}}$, $\hat E_{s,\text{bottom}}$, $\hat E_{s,\text{top}}$), and a separate model is calibrated to each branch. 
    Reported $R^2$ values quantify the agreement of the empirical inverse quadratic fit with the measured branch moduli. 
    a) A representative measured full stress--strain curve, with the characteristic stages of the cyclic test indicated; ``normalized average'' denotes the mean across all trials per sample. 
    b) The measured friction hysteresis loop from the boxed region of (a), annotated with the four branch moduli. 
    c, d) Normalized shear branch moduli versus weaver width: ($\hat E_{s,\text{right}}$, $\hat E_{s,\text{left}}$) in (c) and ($\hat E_{s,\text{bottom}}$, $\hat E_{s,\text{top}}$) in (d). 
    e, f) Normalized shear branch moduli versus weaver spacing: ($\hat E_{s,\text{right}}$, $\hat E_{s,\text{left}}$) in (e) and ($\hat E_{s,\text{bottom}}$, $\hat E_{s,\text{top}}$) in (f). 
    Solid curves are the empirical fits; dashed curves are the corresponding distributed-calibration simulations, one per branch.}
    \label{fig:shearTests}
\end{figure}

\begin{figure}[!t]
    \centering
    \includegraphics[width=0.99\columnwidth]{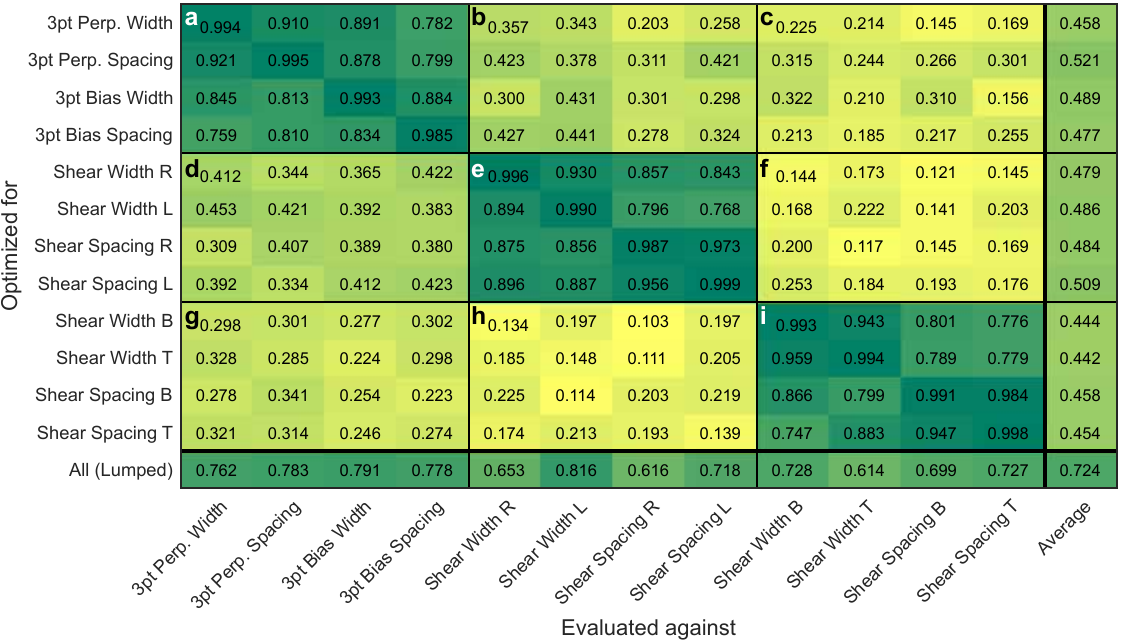}
    \caption{Cross-validation of the calibrated models shows that distributed calibration produces models that generalize poorly across loading scenarios, while lumped calibration yields consistent, moderate performance across all conditions. 
    Each cell reports the $R^2$ of a calibrated model: the $y$-axis indicates the case used for calibration and the $x$-axis the case against which the model was evaluated. 
    The twelve cases fall into three groups: three-point bending, the shear loading/unloading branches (left, right), and the shear plateau branches (top, bottom).
    Within each group, models calibrated to one case predict the others well ($R^2 \approx 0.99$ on the diagonal), but performance degrades sharply across groups ($R^2 < 0.45$), including between different branches of the same hysteresis loop. 
    The lumped calibration (final row) achieves moderate but consistent performance across all cases (average $R^2 = 0.724$). 
    The model can therefore be calibrated for high accuracy on a specific test case or generalized to provide an overall indicator of material response.}
    \label{fig:heatmap}
\end{figure}

\begin{figure}[!t]
    \centering
    \includegraphics[width=0.99\columnwidth]{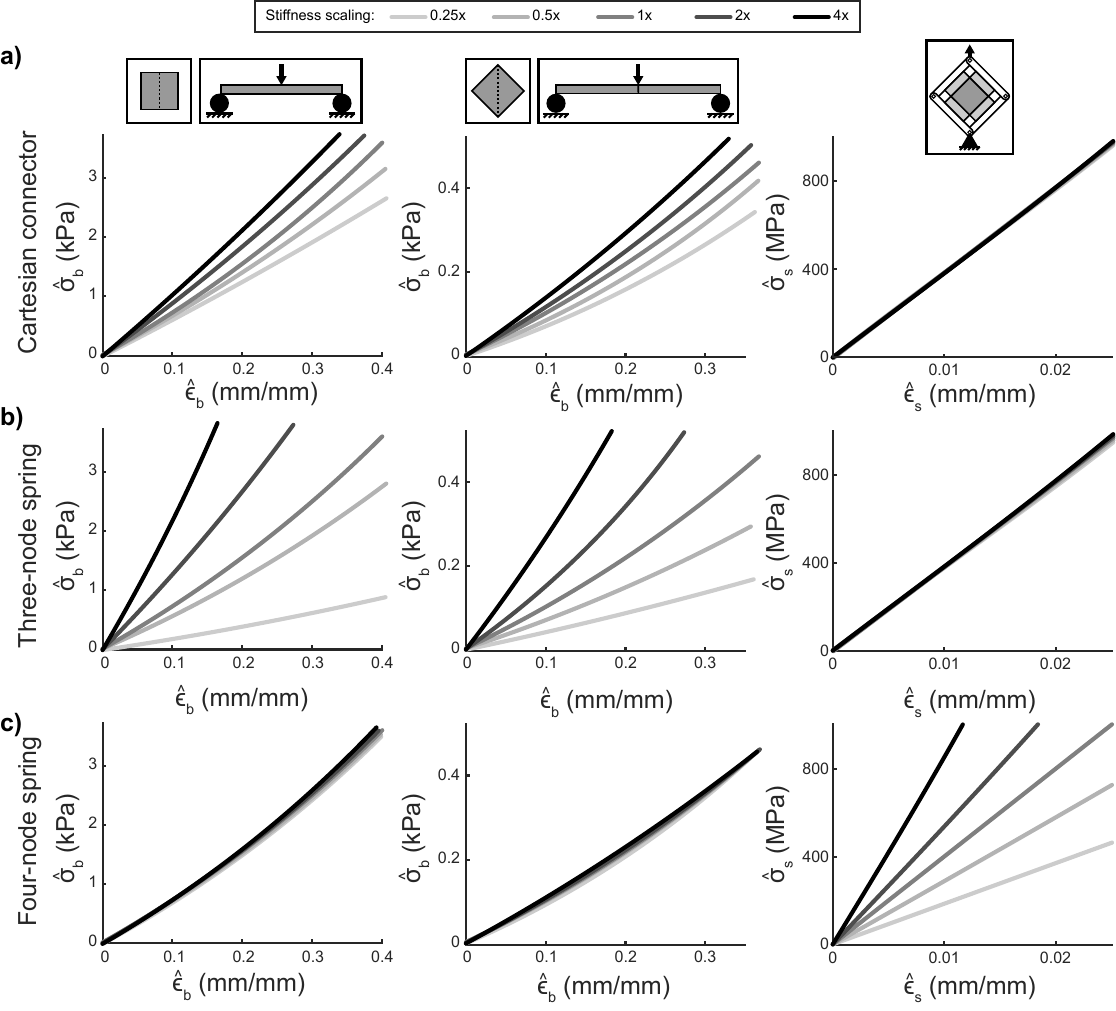}
    \caption{Parametric scaling of individual element stiffnesses reveals the distinct and separable contributions of each element to the bending and shear response. 
    Scaling analyses use the lumped calibration solution as the baseline, with all other element stiffnesses held constant. 
    For shear, the right branch of the hysteresis loop is shown rather than the full loop: the model is single-valued and does not trace a loop, and the complete four-branch treatment is given in Fig.~\ref{fig:shearTests}. 
    The finite slope of this branch is not a frictional property. 
    For an ideal Coulomb-friction system the loop is rectangular (the sliding (top, bottom) branches are perfectly flat and the stuck (left, right) branches perfectly vertical) so an ideal loop has no finite stiffness. 
    The measured branches deviate from this ideal because the weave is elastically compliant; the branch slope therefore reflects the structural elasticity of the weave operating within a given frictional regime.
    The right branch is shown because it corresponds to a stuck regime, in which the weave deforms elastically before slipping and the elastic stiffness is therefore resolvable.
    The top and bottom (sliding) branches carry almost no elastic stiffness and would appear nearly flat under stiffness scaling regardless of the element varied. 
    a) Scaling Cartesian connector stiffness modestly affects the bending response but has negligible effect on shear. 
    b) Scaling three-node spring stiffness significantly affects the bending response but has negligible effect on shear. 
    c) Scaling four-node spring stiffness affects only the shear response, leaving bending unchanged.}
    \label{fig:stiffnessScaling}
\end{figure}

\begin{table}[]
\caption{Stiffness values for distributed and lumped optimization}
\label{tables:opt}
\begin{tabular}{|l|l|l|l|lll|}
\hline
\multicolumn{1}{|c|}{\multirow{2}{*}{\textbf{Optimization Type}}} & \multicolumn{1}{c|}{\multirow{2}{*}{\textbf{Test}}} & \multicolumn{1}{c|}{\multirow{2}{*}{\textbf{Sample set}}} & \multirow{2}{*}{\textbf{Curve portion}} & \multicolumn{3}{c|}{\textbf{Stiffnesses}}                                              \\ \cline{5-7} 
\multicolumn{1}{|c|}{}                                            & \multicolumn{1}{c|}{}                               & \multicolumn{1}{c|}{}                                     &                                         & \multicolumn{1}{c|}{$k_{cart}$} & \multicolumn{1}{c|}{$k_{3spr}$} & \multicolumn{1}{c|}{$k_{4spr}$} \\ \hline
\multirow{12}{*}{Distributed}                                     & \multirow{2}{*}{3pt - Perpendicular}                & Changing widths                                           & All                                     & \multicolumn{1}{l|}{1293}    & \multicolumn{1}{l|}{4574}  & 595                        \\ \cline{3-7} 
                                                                  &                                                     & Changing spacings                                         & All                                     & \multicolumn{1}{l|}{1165}    & \multicolumn{1}{l|}{4622}  & 5628                       \\ \cline{2-7} 
                                                                  & \multirow{2}{*}{3pt - Bias}                         & Changing widths                                           & All                                     & \multicolumn{1}{l|}{955}     & \multicolumn{1}{l|}{4201}  & 6051                       \\ \cline{3-7} 
                                                                  &                                                     & Changing spacings                                         & All                                     & \multicolumn{1}{l|}{1023}    & \multicolumn{1}{l|}{4430}  & 124                        \\ \cline{2-7} 
                                                                  & \multirow{8}{*}{Shear}                              & \multirow{4}{*}{Changing widths}                          & Right                                   & \multicolumn{1}{l|}{788}     & \multicolumn{1}{l|}{106}   & 2010                       \\ \cline{4-7} 
                                                                  &                                                     &                                                           & Left                                    & \multicolumn{1}{l|}{789}     & \multicolumn{1}{l|}{110}   & 1447                       \\ \cline{4-7} 
                                                                  &                                                     &                                                           & Bottom                                  & \multicolumn{1}{l|}{773}     & \multicolumn{1}{l|}{112}   & 15                         \\ \cline{4-7} 
                                                                  &                                                     &                                                           & Top                                     & \multicolumn{1}{l|}{769}     & \multicolumn{1}{l|}{101}   & 8                          \\ \cline{3-7} 
                                                                  &                                                     & \multirow{4}{*}{Changing spacings}                        & Right                                   & \multicolumn{1}{l|}{697}     & \multicolumn{1}{l|}{1159}  & 2047                       \\ \cline{4-7} 
                                                                  &                                                     &                                                           & Left                                    & \multicolumn{1}{l|}{703}     & \multicolumn{1}{l|}{1148}  & 1236                       \\ \cline{4-7} 
                                                                  &                                                     &                                                           & Bottom                                  & \multicolumn{1}{l|}{701}     & \multicolumn{1}{l|}{1153}  & 10                         \\ \cline{4-7} 
                                                                  &                                                     &                                                           & Top                                     & \multicolumn{1}{l|}{689}     & \multicolumn{1}{l|}{1140}  & 9                          \\ \hline
Lumped                                                            & All                                                 & Both                                                      & All                                     & \multicolumn{1}{l|}{1038}    & \multicolumn{1}{l|}{4351}  & 3502                       \\ \hline
\end{tabular}
\end{table}
For three-point bending, the simulation with parameters chosen from the distributed calibration reproduces the empirical normalized bending stiffness across both loading orientations and both geometric parameter sets to within the 1\% error threshold.
The normalized bending stiffness follows the expected scaling with respect to both weaver width and spacing, with the empirical fits to the measured moduli achieving $R^2$ values exceeding 0.92 in all cases (Fig.~\ref{fig:threePointBendTests}). 
The agreement is consistent across the perpendicular and bias orientations, confirming that the model captures the anisotropic bending response of the weave.
For shear loading, the distributed calibrations reproduce stiffnesses extracted from different regions in the hysteresis loop, each monotonic regime fit independently as a piecewise-linear segment, to within their respective thresholds (Fig.~\ref{fig:shearTests}a-b).
The normalized shear stiffness follows the expected scaling with respect to both weaver width and spacing, with the empirical fits achieving $R^2$ values exceeding 0.94 across all cases (Fig.~\ref{fig:shearTests}c-f). 
Because the static model is single-valued, it does not reproduce the hysteresis loop itself; instead, each monotonic branch of the loop is fit independently, so that the loading and unloading branches are represented by separate calibrated stiffnesses rather than by a single continuous frictional response.
The cross-validation heatmap (Fig.~\ref{fig:heatmap}) reveals a clear distinction between the distributed and lumped optimization strategies.
Distributed models achieve near-perfect $R^2$ values within their respective training conditions (regions a, e, and i), but generalize poorly to other loading scenarios, with $R^2$ values dropping below 0.45 outside their training group. 
In contrast, the lumped model achieves consistent $R^2$ values in the approximate range of 0.6 to 0.8 across all conditions, demonstrating that a single set of stiffness values can serve as a reliable general-purpose model of the weave response.
In either case, once the stiffness parameters $\{k_{cart}, k_{3spr}, k_{4spr}\}$ are fixed, the calibrated model predicts the response across weave sizes without recalibration, within the loading regime to which it was calibrated.
The separable contribution of each element to the global response is further illustrated by the parametric stiffness scaling analysis (Fig.~\ref{fig:stiffnessScaling}). 
Scaling the three-node spring stiffness significantly alters the bending response while leaving the shear response unchanged. 
Scaling the four-node spring stiffness affects only the shear response, with no measurable effect on bending. 
Scaling the Cartesian connector stiffness affects the bending response while leaving shear largely unchanged, though its influence on bending is considerably smaller than that of the three-node spring. 
This near-orthogonality between element type and response mode confirms the physical interpretation established in Section~\ref{sec:elementValidation} and provides a principled basis for independent tuning of each element stiffness.

\FloatBarrier
\section{Model Applications}
\label{sec:modelApplications}
The proposed reduced-order model enables direct investigation of how weave-specific mechanical phenomena emerge from local element interactions. 
These behaviors are architecture-driven, arising from the geometric arrangement of the weavers rather than from constituent material properties. 
In the following examples, we demonstrate the ability of the model to reproduce four canonical woven-structure phenomena: crimp interchange, weaver pullout, tearing, and spatially graded stiffness.

\subsection{Crimp Interchange}
\begin{figure}[!t]
    \centering
    \includegraphics[width=0.99\columnwidth]{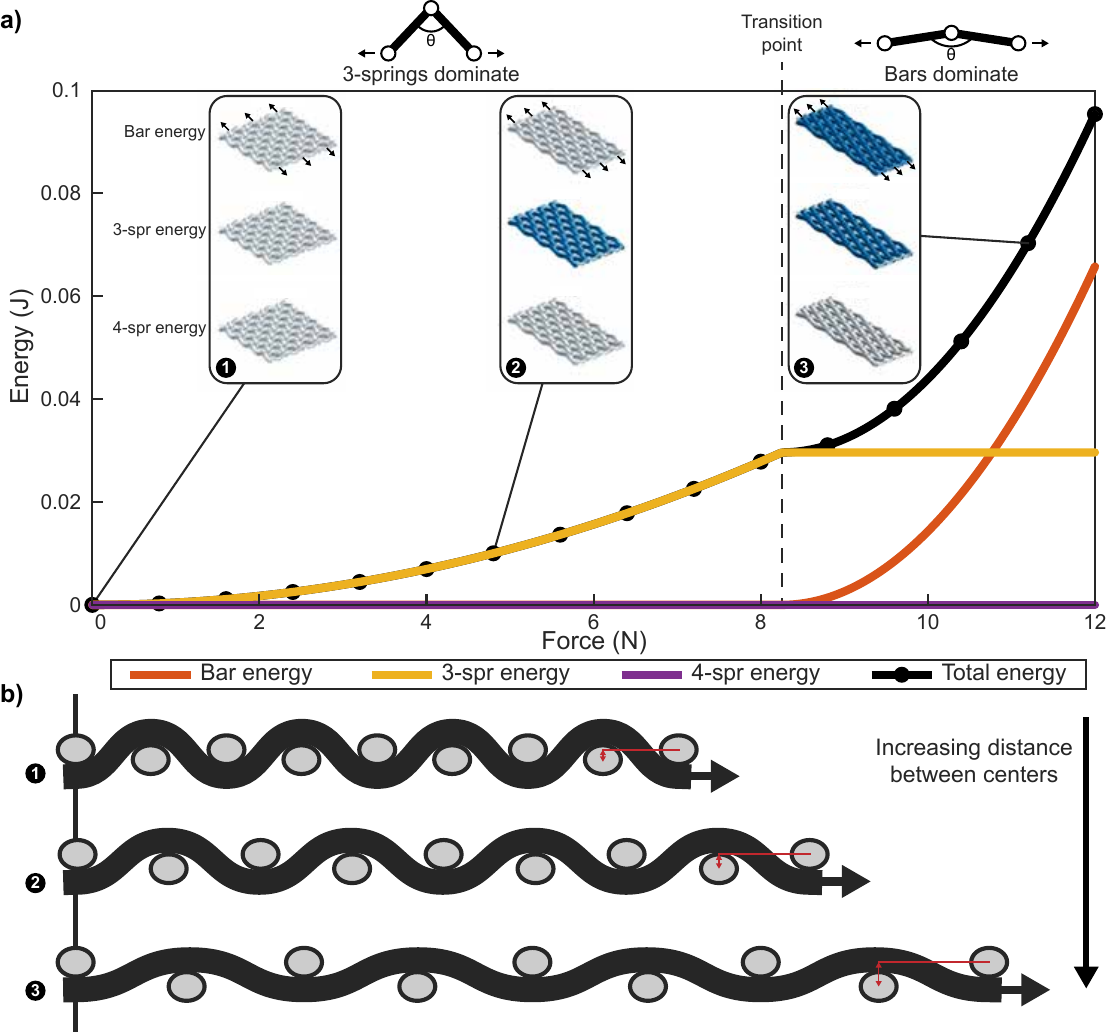}
    \caption{The model captures the emergent Poisson response of a plain woven sheet under uniaxial loading through a transition from crimp-dominated to bar-dominated energy storage. 
    a) Stored energy in each element type as a function of applied force. 
    \emph{This is an energy--force decomposition, not a force--displacement curve; the curve shapes show how stored energy partitions between elements as force increases and should not be read as a stiffness response.} 
    Two regimes are visible: the three-node springs store the energy as the weavers uncrimp and the opening angle $\theta$ increases toward its straightened limit $\theta \to \pi$. 
    As $\theta$ approaches this limit the three-node spring energy saturates and additional energy is stored in axial bar stretching; the dashed line marks this crimp-to-bar transition. 
    The straightened state $\theta = \pi$ is approached asymptotically rather than reached---it would require infinite force---so the top axis maps applied force to the opening angle $\theta$ only up to this limit, and the angle axis terminates as $\theta \to \pi$ rather than continuing. 
    b) Schematic of the deformation mechanism: as a weaver is stretched and straightens, the centers of the orthogonal weavers are driven progressively further apart, coupling the axial extension of one weaver family to the transverse separation of the other.}
    \label{fig:crimpInterchange}
\end{figure}

Crimp interchange is a characteristic behavior of woven materials in which axial tension applied to one set of weavers induces contraction in the orthogonal direction. 
As one set of weavers is pulled taut and straightens, it forces the interlaced orthogonal weavers to take up more crimp and draw closer together, so extension along one axis produces contraction along the other. 
At the global scale, this manifests as an effective Poisson's response, and because it arises purely from the weave geometry, it can occur at small strains without any Poisson's response at the constituent-material level. 
Within the unit cell formulation, this coupling is captured by the three-node springs, whose opening angles link the extension of one weaver to the transverse contraction of the other, and the vertical bars, whose axial compression represents force transferred between the stacked weaver layers at each crossover through their cross-sections.
To demonstrate the effect with our model, one edge of a woven sheet was constrained in two translational DOF while a distributed load was applied along the opposing edge. 
The model reproduces the expected transverse contraction and exhibits a two-regime partitioning of stored energy between the element types (Fig.~\ref{fig:crimpInterchange}). 
At low loads, the stored energy is dominated by the three-node springs as the weavers uncrimp and the opening angle $\theta$ increases toward its straightened limit of $\pi$. 
Near this straightened limit the geometric advantage that converts axial extension into angle change vanishes, so the force required to open the angle further grows sharply.
The three-node spring contribution then saturates and subsequent loading is stored in axial bar stretching instead. 
The straightened state $\theta = \pi$ is an asymptotic limit rather than a configuration the weave attains: it would require infinite force, and in practice the hand-off to bar stretching occurs at a finite angle just short of $\pi$, which we refer to as the transition point. 
This hand-off between energy-storage mechanisms emerges naturally from the element interactions, without any explicit transition law prescribed at the global level.

\subsection{Weaver Pullout}
\begin{figure}[!t]
    \centering
    \includegraphics[width=0.8\columnwidth]{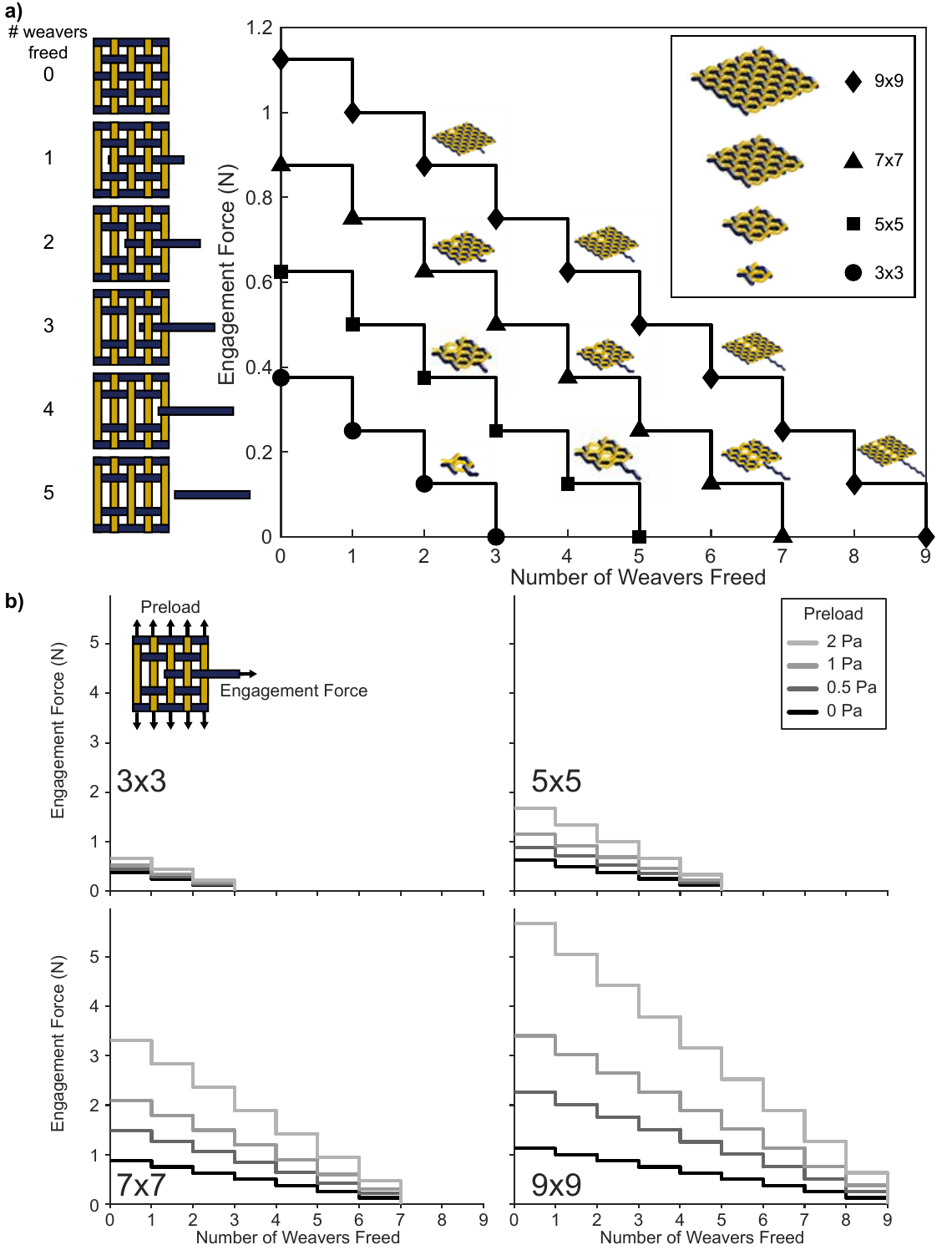}
    \caption{Loss of structural integrity as a weave unravels, evaluated as sets of independent static configurations.
    The model does not simulate frictional pullout directly, instead it predicts the reaction force (engagement force) required to hold a test weaver at a fixed reference displacement.
    a) Engagement force versus number of weavers freed, for $3\times3$ through $9\times9$ weaves under zero preload.
    The force falls in discrete steps as the weave unravels, and larger weaves sustain higher forces because more elements share the load.
    b) Engagement force versus number freed at orthogonal preloads of 0, 0.5, 1, and 2~Pa, for each weave size. 
    Higher preload raises the engagement force, and the amplification is nonlinear: a given increment of preload produces a progressively larger increase in engagement force. 
    This nonlinearity is a mechanical effect of geometric stiffening: preload pre-tensions the orthogonal weavers toward their straightened limit ($\theta \to \pi$), where the engagement stiffness rises steeply, so the same preload increment buys more holding force the closer the weavers already are to straight. 
    It is not an artifact of the preload normalization.
    Preload is applied as a stress specifically so that the orthogonal load per weaver is held consistent across weave sizes, isolating the geometric effect. 
    The amplification grows with weave size because more orthogonal weavers and engagements compound the effect across additional load paths.}
    \label{fig:weaverUnravel}
\end{figure}

Unlike a homogeneous continuum, a woven structure consists of discrete, interlaced weavers held in place by their mutual interlacing. 
As a weave unravels, through fraying, edge wear, or deliberate extraction, weavers become loose one at a time, and the structure progressively loses its ability to hold those that remain. 
In a real fabric, this unraveling is a frictional, rate-dependent sliding process. 
The present model is static and conservative and therefore does not simulate sliding, friction, or any time-dependent process directly.
Instead, it represents the \emph{consequence} of unraveling by re-solving the static problem for successively smaller sets of engaged weavers. 
Each configuration is an independent static solve of a distinct, further-unraveled structure, not a step along a continuous extraction path. 
Each unraveled state is constructed directly: the elements connecting the freed weaver to the rest of the weave are removed, and the freed weaver is truncated so that the unengaged tail does not contribute extraneous elastic effects. 
To quantify how strongly the surrounding weave holds a given weaver, we then prescribe a fixed reference displacement on a single test weaver and record the reaction force required to hold it there, which we call the engagement force. 
The engagement force decreases in discrete steps as successive weavers are freed, because each removal reduces the surrounding structure that resists the prescribed displacement (Fig.~\ref{fig:weaverUnravel}a), and larger weaves sustain higher engagement forces because more elements share the load. 
We further examined the effect of an orthogonal preload across weave sizes of $3\times3$, $5\times5$, $7\times7$, and $9\times9$ (Fig.~\ref{fig:weaverUnravel}b): increasing preload raises the engagement force, and this amplification grows with weave size. 
This amplification arises purely from geometric stiffening, as preload pre-tensions the orthogonal weavers toward their straightened limit where the engagement stiffness rises steeply.
It is elastic in origin and distinct from the frictional amplification that governs preload sensitivity in real woven pullout. 
This analysis carries an important limitation at the final stage of unraveling. 
Because the model represents engagement through persistent elastic elements rather than through frictional contact, it cannot represent the point at which a weaver retains too few engagements to be held at all. 
Physically, once a weaver is freed from all but its last engagement it carries no load and falls free. 
Our model, however, still predicts a finite reaction from the residual elastic elements connecting it. 
The engagement force reported at this last step is therefore a modeling artifact, and we retain it in Fig.~\ref{fig:weaverUnravel} only for completeness.
It should be read as a residual of the elastic formulation rather than a physical holding force. 
A faithful treatment of this regime would require frictional contact and is beyond the current static formulation.

\subsection{Tearing}
\begin{figure}[!t]
    \centering
    \includegraphics[width=0.99\columnwidth]{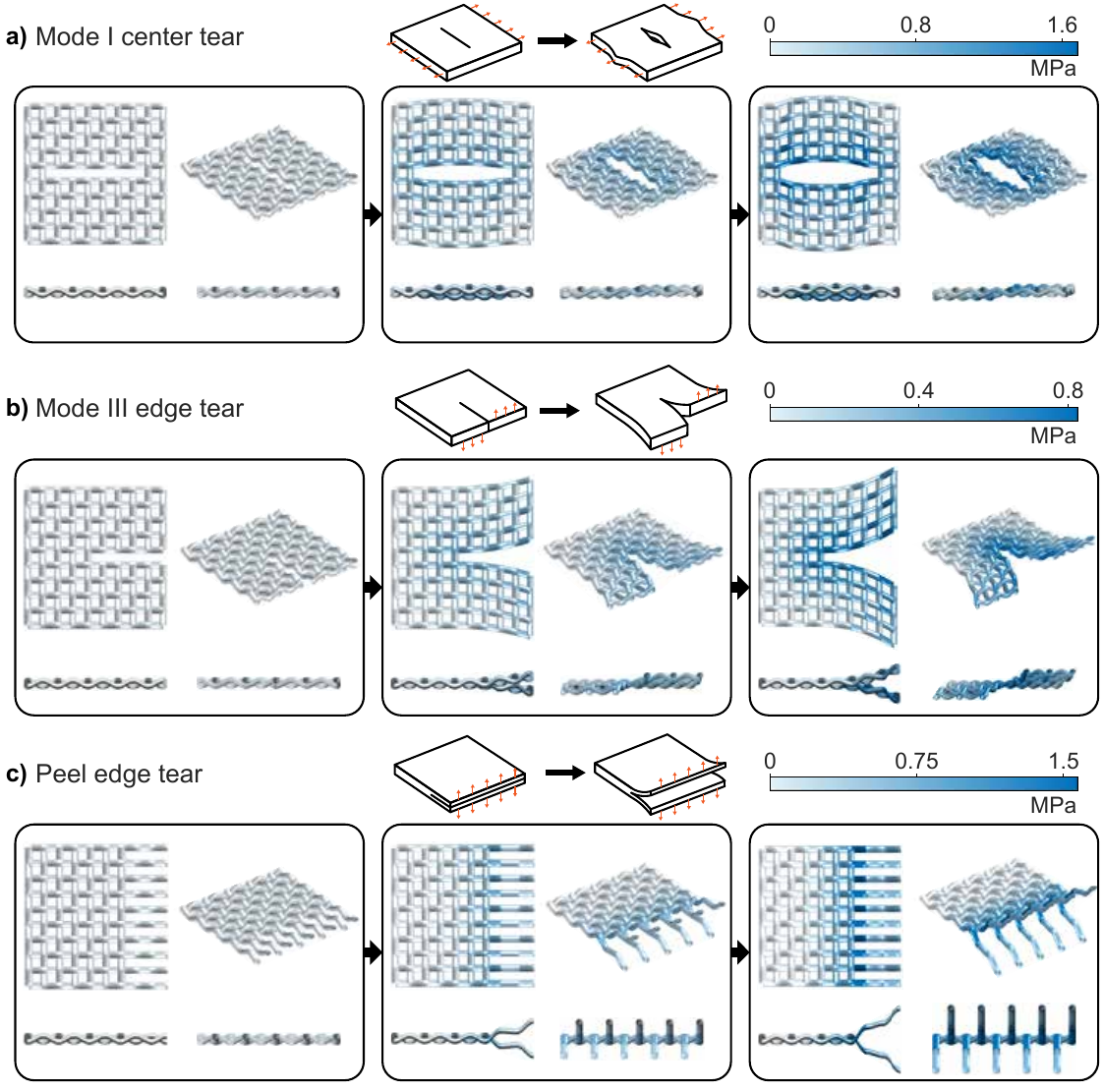}
    \caption{The model resolves element-level stress concentrations under three tearing configurations, identifying likely damage-initiation sites in woven sheets. 
    For each configuration, one edge of a sheet containing an initial crack is fully constrained while prescribed loads open the crack. 
    Color indicates stress magnitude (MPa).
    Note that the stress scale differs between panels. 
    a) Mode~I center tear: equal and opposite in-plane tensile loads across a centered crack drive symmetric opening, concentrating stress at the crack tips. 
    b) Mode~III edge tear: equal and opposite out-of-plane loads across a cracked edge produce tearing-mode deformation localized at the crack tip. 
    c) Peel configuration: one layer is progressively separated from the other along a free edge, concentrating stress at the peel front. 
    The model is linear-elastic and does not simulate crack growth.
    The stress fields indicate where damage would initiate, not a computed propagation path.}
    \label{fig:tearing}
\end{figure}

While woven structures share the tendency to develop stress concentrations around discontinuities such as holes or tears with homogeneous solids, the discrete nature of the weave introduces additional complexity in how damage localizes. 
The element-wise formulation of the present model resolves stress distributions at the individual weaver level, enabling direct identification of regions susceptible to damage. 
We emphasize that the model is linear-elastic and contains no failure criterion or crack-advance mechanism.
It merely computes the stress field around a fixed, pre-existing crack and therefore identifies likely damage-\emph{initiation} sites rather than simulating crack growth. 
Three tearing configurations were simulated, each with one edge of the sheet fully constrained and prescribed loads applied to open the crack (Fig.~\ref{fig:tearing}). 
In the Mode~I center tear, equal and opposite in-plane tensile loads were applied across a centered crack, producing symmetric opening with stress concentrated at the crack tips. 
In the Mode~III edge tear, equal and opposite out-of-plane loads were applied across a cracked edge, producing tearing-mode (out-of-plane shear) deformation localized at the crack tip. 
In the peel configuration, one layer was progressively separated from the other along a free edge, concentrating stress at the advancing peel front. 
These three configurations span an opening mode (Mode~I), an out-of-plane shear or tearing mode (Mode~III), and a delamination-type peel; the in-plane shear mode (Mode~II) is outside the scope of the present demonstration. 
Across all three cases, the element-level stress field identifies the weavers carrying the highest load, indicating where damage would be most likely to initiate.

\subsection{Variable Stiffness}
\begin{figure}[!t]
    \centering
    \includegraphics[width=0.99\columnwidth]{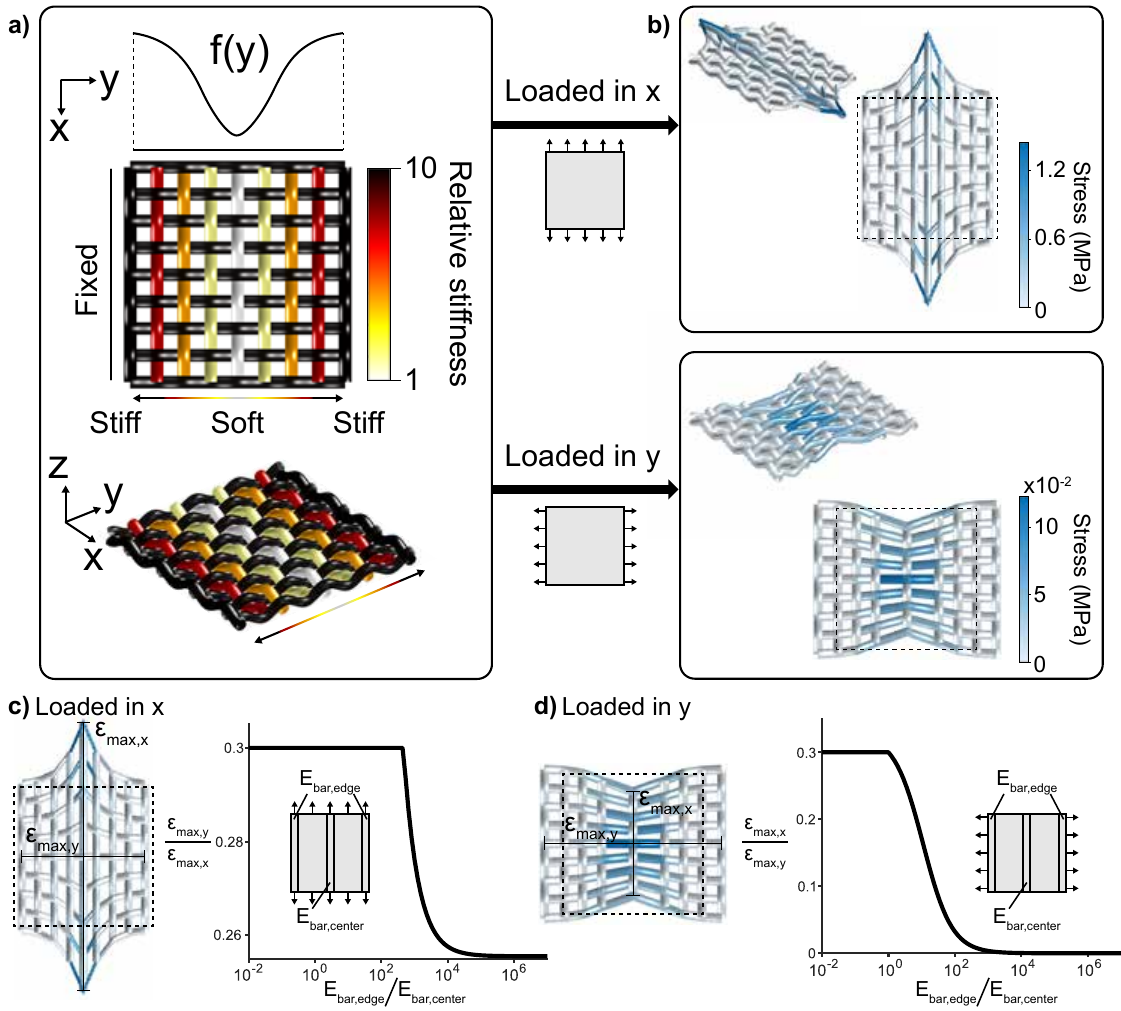}
    \caption{Element-wise stiffness parameterization enables spatially graded woven structures with programmable, direction-dependent anisotropy. 
    a) A unidirectional stiffness gradient is applied along $y$ (stiff edges, compliant center) while stiffness along $x$ is uniform.
    The relative-stiffness gradient shown spans a factor of ten. 
    b) Deformed configurations and element stress under $x$- and $y$-directed loading. 
    Loading along $x$ localizes deformation in the compliant center while loading along $y$ deforms more uniformly. 
    Note the stress color scales differ between the two cases by roughly an order of magnitude.
    c, d) Effective Poisson response under $x$- and $y$-directed loading, respectively.
    In each panel the deformed configuration (left) is annotated with the maximum spanwise strains $\varepsilon_{max,x}$ and $\varepsilon_{max,y}$, and the plot (right) reports the transverse-to-axial strain ratio against the bar-modulus gradient $E_{bar,edge}/E_{bar,center}$ (edge modulus fixed, so larger values are a stronger stiff-edge, soft-center gradient).
    The transverse-to-axial ratio is $\varepsilon_{max,y}/\varepsilon_{max,x}$ in (c) and $\varepsilon_{max,x}/\varepsilon_{max,y}$ in (d).
    The two vertical axes use different scales: (c) spans $0.255$--$0.3$ while (d) spans $0$--$0.3$, reflecting the much larger change under $y$ loading.}
    \label{fig:variableStiffness}
\end{figure}

Conventional homogenized models of woven structures assign a single set of effective properties to the entire structure, making them unable to capture spatially varying or weaver-level stiffness distributions. 
In contrast, the proposed model assigns stiffness parameters independently to each element. 
Because each weaver is discretized into multiple elements, stiffness can vary from weaver to weaver and along the length of a single weaver, enabling continuously graded or locally heterogeneous architectures. 
To illustrate this, a unidirectional stiffness gradient was applied across the $y$-direction, with weavers transitioning from stiff at the edges to compliant at the center, while stiffness along $x$ remained uniform (Fig.~\ref{fig:variableStiffness}a). 
The gradient is imposed through the bar modulus alone: the edge weavers are assigned a fixed modulus $E_{bar,edge}$ and the central weavers a smaller modulus $E_{bar,center}$, so the dimensionless ratio $E_{bar,edge}/E_{bar,center}$ sets the strength of the gradient, with $E_{bar,edge}/E_{bar,center}=1$ recovering a uniform sheet and larger values producing a progressively stiffer edge relative to the center.
The sheet was then loaded independently in $x$ and $y$, and the resulting deformed configurations and stress distributions compared (Fig.~\ref{fig:variableStiffness}b). 
Loading along $x$ concentrates tensile deformation in the compliant central weavers, producing a pronounced hourglass extension.
Loading along $y$ distributes deformation more evenly but draws the central region inward, as the highly stretched compliant weavers pull their orthogonal neighbors together through crimp interchange.
The two cases also differ considerably in stress magnitude, by roughly an order of magnitude.
To quantify the resulting anisotropy, we define the maximum spanwise strains $\varepsilon_{max,x}$ and $\varepsilon_{max,y}$ as the peak strains measured along $x$ and $y$ at full load, and for each loading direction track the transverse-to-axial strain ratio against the gradient ratio $E_{bar,edge}/E_{bar,center}$ (Fig.~\ref{fig:variableStiffness}c,d).
Under both loading directions the ratio begins near $0.3$ when the gradient is weak and then changes as the gradient strengthens, plateauing beyond a ratio of approximately $10^2$--$10^3$ once the stiffness contrast is large enough that the response saturates.
The two directions plateau at distinct values: $y$ loading, aligned with the gradient, drives the ratio toward zero, whereas $x$ loading, across the uniform direction, changes it only modestly. It is this difference between the two plateaus that constitutes the programmed anisotropy.
These results demonstrate that targeted mechanical anisotropy can be programmed into a woven structure through weaver-level stiffness tuning alone, without modifying the weave architecture or constituent material.

\FloatBarrier
\section{Discussion}
\label{sec:discussion}
The results presented in this work demonstrate that a reduced-order, element-based framework can reliably reproduce the weave-specific mechanical phenomena that emerge from the geometric arrangement of interlaced weavers. 
By explicitly representing fundamental deformation mechanisms at the element level and assembling them into repeating unit cells, the model bridges the gap between continuum homogenization approaches (which cannot capture these phenomena) and high-fidelity finite element simulations (which resolve them at prohibitive computational cost). 
The following sections discuss the capabilities enabled by this formulation and its inherent limitations.

\subsection{Model Capabilities}
A primary strength of the proposed model is its ability to capture architecture-driven emergent behaviors that are inaccessible to homogenized continuum descriptions. 
Mechanical responses, such as anisotropic stiffness, crimp interchange, shear-induced locking, and localized deformation modes, arise naturally from the geometric arrangement of the weavers and their interactions, rather than being imposed through phenomenological constitutive laws. 
This is significant because it means the model remains scale-independent and applicable across a wide range of woven architectures without re-parameterization, provided the underlying material behavior is approximately linear elastic. 
In this respect the model extends the bar-and-hinge lineage of reduced-order structural models, originally developed for origami and recently applied to ribbons and woven structures \citep{Schenk2011,filipov2017bar,Liu2017,zhu2025lumped}, to the specific interaction modes of woven architectures. 
Like those formulations, it derives its behavior from a small set of analytically defined elements rather than a fine continuum mesh; unlike them, its element stiffnesses are calibrated to capture the weaver-scale in-plane and crimp couplings that govern weave-specific response. 

The modular element-based formulation enables direct physical interpretation of deformation mechanisms. 
Each stiffness element corresponds to a distinct mode of interaction, allowing individual contributions to the global response to be isolated, validated, and tuned independently. 
The unit-cell eigenmode analysis confirms that the full complement of elements is necessary to reproduce the expected elastic modes of woven systems, and the stiffness scaling analysis demonstrates that each element contributes orthogonally to the global response. 
Together, these properties make the framework transparent and physically interpretable in a way that black-box finite element models are not. 

From a computational standpoint, the reduced-order nature of the model offers significant efficiency gains over traditional finite element analysis. 
Each node carries only three translational degrees of freedom and each unit cell only a handful of analytically defined elements, so the global system remains compact even for very large domains; this minimal-DOF construction is the same source of efficiency that makes bar-and-hinge models far cheaper than continuum discretizations \citep{zhu2025lumped}. 
This makes the framework well suited for parametric studies, stiffness calibration, and design space exploration, and is particularly relevant for applications such as inverse design of woven architected materials, where large numbers of repeated simulations are required. 

The unit-cell construction also facilitates modeling of spatially graded or heterogeneous weaves. 
By adjusting element stiffnesses locally, at the level of individual weavers or even individual elements within a weaver, the model can represent functionally graded architectures and explore variable stiffness, tailored deformation pathways, and directional anisotropy. 

\subsection{Model Limitations}
Despite its advantages, the proposed model has several limitations that should be acknowledged. 
The most fundamental is that its element stiffnesses are effective parameters, rather than quantities derived from first principles. 
Because the model does not compute stiffness from constituent geometry and material properties, it cannot make quantitative predictions until it has been calibrated against experimental data, and this calibration is specific to the material and weave system used. 
A parameter set calibrated for one material does not transfer to another: changing the weaver material, cross-section, or fabrication would, in general, require the calibration procedure to be repeated. 
The calibration step is therefore a recurring prerequisite rather than a one-time cost, and it is the principal trade-off accepted in exchange for the efficiency and physical transparency of the model. 
A model that instead derived element stiffnesses directly from constituent properties would avoid this step, and developing such a closure is a natural direction for future work. 

The formulation also assumes linear elastic material behavior and does not account for material nonlinearity, plasticity, or viscoelastic effects. 
While weave-specific phenomena are dominated by geometric interactions, material nonlinearities may become significant in applications involving large strains or damage accumulation, and the current framework would need to be extended to address such cases. 
Contact interactions between weavers are represented using simplified stiffness elements that capture slip and pseudo-friction rather than fully resolved contact mechanics. 
As a result, the model does not explicitly enforce contact constraints or capture detailed stress distributions at contact interfaces. 
Phenomena such as frictional hysteresis or stick-slip transitions are therefore approximated rather than resolved rigorously, and applications requiring precise contact force predictions may require augmentation with dedicated contact formulations. 
The centerline-based nodal representation limits the ability of the model to capture cross-sectional deformation, local buckling, or three-dimensional stress states within individual weavers. 
For applications where local stress states or failure initiation are of primary interest, higher-fidelity finite element models remain necessary. 
The present framework is therefore best understood as complementary to, rather than a replacement for, high-fidelity simulation approaches. 
Finally, the unit-cell tessellation is well suited to periodic and quasi-periodic woven architectures but may require additional considerations for highly irregular weaves or structures with significant topological defects. 

Extensions to the current framework, such as adaptive discretization or enhanced contact formulations, could broaden its applicability further. 
More broadly, the demonstrated ability to reproduce weave-specific phenomena efficiently and with physical transparency positions this framework as a practical tool for the design of woven architected materials, supporting the exploration of programmable anisotropy, multifunctional response, and geometry-driven mechanical behavior across scales and applications.
\FloatBarrier
\section{Conclusion}
\label{sec:conclusion}

This work presents a reduced-order modeling framework for woven structures that captures weave-specific mechanical phenomena arising from the geometric arrangement of interconnected weavers. 
By representing woven systems as tessellations of repeating unit cells composed of physically interpretable stiffness elements, the model occupies a practical middle ground between homogenized continuum descriptions, which cannot capture these phenomena, and high-fidelity finite element simulations, which resolve them at substantial computational cost.

The framework was validated at multiple levels: element-level testing against analytical solutions confirms correct implementation of each deformation mechanism, unit-cell eigenmode analysis confirms that the combined formulation captures the full expected elastic behavior of woven systems, and empirical calibration demonstrates quantitative agreement with physical samples under bending and shear loading. 
The model applications further demonstrate that crimp interchange, weaver pullout, tearing, and spatially graded stiffness all emerge naturally from the geometric arrangement of the weavers without phenomenological 
constitutive assumptions, confirming the physical basis of the formulation.

Taken together, these results establish the proposed framework as a computationally efficient and physically transparent tool for the analysis and design of woven structures. 
The ability to assign stiffness parameters independently at the element level opens a direct pathway for programming mechanical anisotropy, tailoring deformation responses, and exploring the design space of woven architected materials. 
This positions weaving not merely as a manufacturing method but as a geometry-driven design strategy for achieving programmable mechanical behavior across a broad range of structural applications.
\FloatBarrier
\section*{Acknowledgments}
\label{sec:acknowledgments}

We acknowledge support from National Science Foundation (NSF) in the form of a NSF Graduate Research Fellowship awarded to Anvay A. Pradhan and support from the Air Force Office of Scientific Research (AFOSR) in the form of a AFOSR grant (AWD021427 FA9550-22-1-0321) awarded to Dr. Evgueni T. Filipov.
\FloatBarrier
\section*{Code and data availability}
The model implementation and the data supporting the findings of this study are not publicly hosted but are available from the corresponding author upon reasonable request.

\FloatBarrier

\appendix


\section*{Appendix A: Computation of derivatives}
\label{sec:AppendixA}

Every element in the reduced-order model is defined by a scalar strain energy $\Pi$ that depends only on the nodal coordinates. Once $\Pi$ is written down, the two quantities the solver needs follow by differentiation. The internal force vector is the gradient $\partial\Pi/\partial\underline{u}$, and the element tangent stiffness is the Hessian $\partial^2\Pi/\partial\underline{u}^2$. This is the standard potential-energy route to the matrix structural analysis of bar-and-hinge origami models \citep{Schenk2011,Liu2017,filipov2017bar}, applied here to the four element types of the weave: the Cartesian connector, the bar, the three-node spring, and the four-node spring.

Because each energy is a function of a single geometric measure, a length $L$ for the connector and bar or a dihedral angle $\theta$ for the springs, the chain rule separates each derivative cleanly into two contributions. The first is a material term, the outer product of the first derivative of the measure with itself, which is always present. The second is a geometric term, proportional to the current force or moment times the second derivative of the measure, which captures the stiffening or softening that arises from member reorientation under load. At the undeformed reference state the force or moment is zero, so the geometric term vanishes and only the material term survives. This material term is exactly the linear element stiffness used by the small-displacement assembly, while the full nonlinear tangent retains both contributions. The same two-term decomposition appears in the nonlinear bar-and-hinge formulations of \citet{Liu2017} and \citet{filipov2017bar}, and the bar derivation that follows is adapted from the treatment of geometrically nonlinear truss elements in McGuire, Gallagher, and Ziemian \citeyearpar{mcguire2000}.

The subsections below give the analytical derivatives for each element. Section~A.2 then documents the finite-difference scheme used to verify them.

\subsection*{A.1. Analytical computation of derivatives}

\subsubsection*{A.1.1 Cartesian connectors}

The Cartesian connector is the simplest element. It is a one-dimensional spring that couples a single coordinate direction of two nodes. In the weave it ties the two stacked weaver layers together where they share an $(x,y)$ location, along a direction in which an ordinary $E A/L_0$ bar would have a zero reference length and divide by zero. Its energy is the quadratic penalty on the change in the signed scalar coordinate separation $L$,
\begin{align}
    \Pi_{cart} = \frac{1}{2}k_{cart}\left(L-L_0\right)^2 \in \mathbb{R}^{1\times1} .
\end{align}

Differentiating once gives the internal force and twice gives the stiffness, following the material and geometric split described in the appendix introduction,
\begin{align}
    \partial \Pi_{cart} = k_{cart}\left(L-L_0\right)\partial L \in \mathbb{R}^{2\times1} ,
\end{align}
\begin{align}
    \partial^2 \Pi_{cart} = k_{cart}\left(L-L_0\right)\partial^2 L + k_{cart}\,\partial L\,(\partial L)^T \in \mathbb{R}^{2\times2} .
\end{align}

Because the connector acts along a fixed coordinate axis, $L$ is the difference of one coordinate of the two nodes and is therefore a linear function of the displacements. The reference and current separations are
\begin{align}
    L_0 = \|x_{ij}-x_{mn}\| = x_{ij}-x_{mn} ,
\end{align}
\begin{align}
    L = \|x_{ij}'-x_{mn}'\| = x_{ij}'-x_{mn}' ,
\end{align}
where the norm collapses to a signed difference under the convention that the ordering of the two coupled degrees of freedom is fixed, so that $x_{ij}>x_{mn}$. The first derivative with respect to the two coupled degrees of freedom is therefore constant,
\begin{align}
    \partial L =
    \begin{bmatrix}
        -1 \\
        1
    \end{bmatrix} ,
\end{align}
and because $L$ is linear its second derivative vanishes identically,
\begin{align}
    \partial^2 L =
    \begin{bmatrix}
        0 & 0 \\
        0 & 0
    \end{bmatrix} .
\end{align}

The vanishing of $\partial^2 L$ means the geometric term in $\partial^2\Pi_{cart}$ drops out for all configurations rather than only at the reference state. The connector stiffness is therefore the constant rank-one matrix $k_{cart}\,\partial L\,(\partial L)^T = k_{cart}\!\left[\begin{smallmatrix}1&-1\\-1&1\end{smallmatrix}\right]$, which is the stiffness of a linear axial spring.

In the code this stiffness is supplied directly as a single connector constant \texttt{k\_cart}, so the element is assembled as a pure $2\times2$ spring. The constant $k_{cart}$ plays the role of the $E A/L_0$ factor of an ordinary axial bar, but supplying it directly avoids the $1/L_0$ that would be singular for the coincident layers the connector ties together, which is what motivated introducing the connector in the first place.

\subsubsection*{A.1.2 Bar elements}

The bar element is a standard two-node axial truss that carries the in-plane stretching stiffness of the weavers. Its strain energy penalizes the change in member length between the reference length $L_0$ and the current length $L$,
\begin{align}
    \Pi_{bar} = \frac{1}{2}\frac{E_{bar}A_{bar}}{L_0}\left(L-L_0\right)^2 \in \mathbb{R}^{1\times1} .
\end{align}

This is the classical nonlinear bar. The derivation below follows the matrix structural analysis treatment of geometrically nonlinear truss elements in McGuire, Gallagher, and Ziemian \citeyearpar{mcguire2000} and is the same element used in the nonlinear bar-and-hinge models of \citet{Liu2017} and \citet{filipov2017bar}. As before, one differentiation gives the internal force and two give the tangent stiffness, with the geometric term carrying the axial-force contribution,
\begin{align}
    \partial \Pi_{bar} = \frac{E_{bar}A_{bar}}{L_0}\left(L-L_0\right)\partial L \in \mathbb{R}^{6\times1} ,
\end{align}
\begin{align}
    \partial^2 \Pi_{bar} = \frac{E_{bar}A_{bar}}{L_0}\left(L-L_0\right)\partial^2 L + \frac{E_{bar}A_{bar}}{L_0}\,\partial L\,(\partial L)^T \in \mathbb{R}^{6\times6} .
\end{align}
The material term $\frac{E_{bar}A_{bar}}{L_0}\,\partial L(\partial L)^T$ evaluates to $\frac{E_{bar}A_{bar}}{L_0}\,\underline{n}\,\underline{n}^T$ in each $3\times3$ block, where $\underline{n}$ is the axial unit vector, and is the linear stiffness used at the reference state. The geometric term contributes $\frac{N}{L}(I-\underline{n}\,\underline{n}^T)$ blocks with axial force $N=\frac{E_{bar}A_{bar}}{L_0}(L-L_0)$. It accounts for transverse reorientation of the loaded member and vanishes when $L=L_0$.

Unlike the connector, the bar length is a nonlinear function of the displacements, so $\partial^2 L\neq0$ and the chain rule is carried out in full. Let $\underline{r} = \underline{x}_i' - \underline{x}_j'$ denote the current chord vector, with reference and current lengths
\begin{align}
    L_0 = \|\underline{x}_i - \underline{x}_j\| , \qquad
    L = \|\underline{r}\| = \sqrt{\underline{r}^T \underline{r}} .
\end{align}
Writing $\phi = \underline{r}^T\underline{r} = L^2$, the derivative of $L=\phi^{1/2}$ is
\begin{align}
    \partial L = \tfrac{1}{2}\phi^{-1/2}\,\partial \phi ,
\end{align}
and the second derivative applies the product rule to $\tfrac{1}{2}\phi^{-1/2}\partial\phi$, which produces one term in $\partial^2\phi$ and one outer-product term through the $-\tfrac{1}{4}\phi^{-3/2}$ derivative of $\phi^{-1/2}$,
\begin{align}
    \partial^2 L = \tfrac{1}{2}\phi^{-1/2}\,\partial^2 \phi - \tfrac{1}{4}\phi^{-3/2}\,\partial\phi\,(\partial\phi)^T .
\end{align}
Here $\partial\phi$ is taken as a column gradient, so that $\partial\phi\,(\partial\phi)^T$ is the $6\times6$ outer product.

The inner quantity $\phi$ is quadratic in the coordinates, so its own derivatives are elementary. Its gradient is
\begin{align}
    \partial \phi = 2\,(\partial \underline{r})^T \underline{r} ,
\end{align}
and its Hessian, by the product rule and the linearity of $\underline{r}$, is
\begin{align}
    \partial^2 \phi = 2\,(\partial \underline{r})^T \partial\underline{r} .
\end{align}

Finally, the chord vector $\underline{r}$ is linear in the six nodal degrees of freedom, so its first derivative is the constant signed selector and its second derivative is zero,
\begin{align}
    \partial \underline{r} =
    \begin{bmatrix}
        I_{3\times3} & -I_{3\times3}
    \end{bmatrix} , \qquad
    \partial^2 \underline{r} = 0 .
\end{align}
Substituting back up the chain collapses these into the compact $\underline{n}\,\underline{n}^T$ material block and $(I-\underline{n}\,\underline{n}^T)$ geometric block structure quoted above.

The linear bar matrix (\texttt{computeKlocal\_bar}) uses only the material block $\frac{E A}{L_0}\,\underline{n}\,\underline{n}^T$ since the reference axial force is zero. The nonlinear element (\texttt{computeFK\_bar}) adds the geometric block $\frac{N}{L}(I-\underline{n}\,\underline{n}^T)$ exactly as above. If a hyperelastic bar is desired in place of the engineering $E A/L_0$ form, the same two-term structure holds with $E A/L_0$ replaced by the tangent modulus times area and the corresponding axial stress, as in the nonlinear bar-and-hinge formulations cited above.

\subsubsection*{A.1.3 Three-node springs}

The three-node spring penalizes the in-plane opening angle $\theta$ at a middle node between its two incident weaver segments, supplying the uncrimping stiffness. Its energy is a quadratic penalty on the deviation of $\theta$ from its reference value $\theta_0$,
\begin{align}
    \Pi_{3spr} = \frac{1}{2}k_{3spr}(\theta-\theta_0)^2 \in \mathbb{R}^{1\times1} .
\end{align}
The force and stiffness follow the material and geometric split, now with the opening angle $\theta$ playing the role of the geometric measure,
\begin{align}
    \partial \Pi_{3spr} = k_{3spr}(\theta-\theta_0)\,\partial\theta \in \mathbb{R}^{9\times1} ,
\end{align}
\begin{align}
    \partial^2 \Pi_{3spr} = k_{3spr}(\partial\theta)(\partial\theta)^T + k_{3spr}(\theta-\theta_0)\,\partial^2\theta \in \mathbb{R}^{9\times9} .
\end{align}
The first term $k_{3spr}(\partial\theta)(\partial\theta)^T$ is the material stiffness. The second term $k_{3spr}(\theta-\theta_0)\partial^2\theta$ is the geometric term and vanishes at the reference state $\theta=\theta_0$.

The difficulty of a rotational spring is differentiating the angle robustly. Defining $\theta$ through $\arccos$ or $\arcsin$ introduces a $1/\sin\theta$ or $1/\cos\theta$ factor in the gradient that grows without bound near the flat and folded configurations \citep{Liu2017,filipov2017bar}. Following \citet{Liu2017}, we instead build $\theta$ from the cross product $d=\|a\times b\|$ and the inner product $c=a\cdot b$ and take $\theta=\operatorname{atan2}(d,c)$, which is single-valued over the full range and whose derivatives stay finite away from the exactly degenerate points. The reference and current angles are
\begin{align}
    \theta_0 = \operatorname{atan2}\!\left(\|\underline{v}_1 \times \underline{v}_2\|,\ \underline{v}_1 \cdot \underline{v}_2\right) , \qquad
    \theta = \operatorname{atan2}(d,c) ,
\end{align}
with reference segment vectors
\begin{align}
    \underline{v}_1 = \underline{x}_1 - \underline{x}_2 , \qquad
    \underline{v}_2 = \underline{x}_3 - \underline{x}_2 ,
\end{align}
and their deformed counterparts $a$ and $b$,
\begin{align}
    a = (\underline{x}_1+\underline{u}_1) - (\underline{x}_2+\underline{u}_2) , \qquad
    b = (\underline{x}_3+\underline{u}_3) - (\underline{x}_2+\underline{u}_2) ,
\end{align}
from which the inner and cross product quantities follow,
\begin{align}
    c = a^T b , \qquad s = a \times b , \qquad d = \|s\| .
\end{align}

Because $a$ and $b$ are linear in the nodal displacements, the nodal gradient assembles from the two edge-space gradients $\partial_a\theta$ and $\partial_b\theta$ through the constant chain $\partial a/\partial u_1 = I$, $\partial a/\partial u_2 = -I$, $\partial b/\partial u_2 = -I$, and $\partial b/\partial u_3 = I$,
\begin{align}
    \partial \theta =
    \begin{bmatrix}
    \partial_{\underline{u}_1}\theta \\
    \partial_{\underline{u}_2}\theta \\
    \partial_{\underline{u}_3}\theta
    \end{bmatrix}
    =
    \begin{bmatrix}
    \partial_a\theta \\
    -\partial_a\theta - \partial_b\theta \\
    \partial_b\theta
    \end{bmatrix} .
\end{align}
Differentiating $\theta=\operatorname{atan2}(d,c)$ gives the standard form
\begin{align}
    \partial_a\theta = \frac{c\,\partial_a d - d\,\partial_a c}{L} , \qquad
    \partial_b\theta = \frac{c\,\partial_b d - d\,\partial_b c}{L} ,
\end{align}
where the denominator $L=c^2+d^2$ simplifies. Using $d^2=\|a\times b\|^2 = \|a\|^2\|b\|^2 - (a^Tb)^2$, the $c^2$ terms cancel and
\begin{align}
    L = c^2 + d^2 = c^2 + \left(\|a\|^2\|b\|^2 - c^2\right) = \|a\|^2\|b\|^2 .
\end{align}

The pieces of the gradient are the inner and cross product derivatives. The inner product derivatives are immediate from $c=a^Tb$,
\begin{align}
    \partial_a c = b , \qquad \partial_b c = a .
\end{align}
The cross product magnitude $d=\|s\|$ differentiates through its unit vector $n=s/\|s\|$,
\begin{align}
    \partial d = \frac{s^T}{\|s\|}\,\partial s = n^T \partial s , \qquad n = \frac{s}{\|s\|} ,
\end{align}
and the cross product itself is written with the skew-symmetric matrix $[\,\cdot\,]_\times$, which turns a cross product into a linear map,
\begin{align}
    s = a \times b = [a]_\times b = -[b]_\times a , \qquad
    [a]_\times =
    \begin{bmatrix}
    0 & -a_3 & a_2 \\
    a_3 & 0 & -a_1 \\
    -a_2 & a_1 & 0
    \end{bmatrix} .
\end{align}
Holding one edge fixed at a time gives the per-edge cross product derivatives,
\begin{align}
    \partial_a s = -[b]_\times , \qquad \partial_b s = [a]_\times ,
\end{align}
and hence, through $\partial d = n^T\partial s$ and the skew identity $n^T[b]_\times = -([b]_\times n)^T$,
\begin{align}
    \partial_a d = [b]_\times n , \qquad \partial_b d = -[a]_\times n .
\end{align}
These are written as column-vector gradients, with the leading transpose suppressed by the gradient-as-column convention.

The nodal Hessian is assembled from the edge-space block Hessian, mapped to the nine nodal degrees of freedom by the constant chain matrix $T$,
\begin{align}
    H_{\underline{u}}\theta = T^T H_{ab}\, T , \qquad
    T =
    \begin{bmatrix}
    I_{3\times3} & -I_{3\times3} & 0_{3\times3} \\
    0_{3\times3} & -I_{3\times3} & I_{3\times3}
    \end{bmatrix} ,
\end{align}
\begin{align}
    H_{ab} =
    \begin{bmatrix}
    \partial_a\partial_a\theta & \partial_a\partial_b\theta \\
    \partial_b\partial_a\theta & \partial_b\partial_b\theta
    \end{bmatrix} .
\end{align}
Each block is the derivative of the corresponding gradient $\partial_a\theta = t_a/L$ or $\partial_b\theta = t_b/L$, whose numerators are
\begin{align}
    t_a = c\,\partial_a d - d\,\partial_a c = c[b]_\times n - d\,b , \qquad
    t_b = c\,\partial_b d - d\,\partial_b c = -c[a]_\times n - d\,a ,
\end{align}
so that by the quotient rule each block has the form $\frac{1}{L}\partial t - \frac{t}{L^2}\partial L$. The $aa$ block is
\begin{align}
    \partial_a\partial_a\theta = \frac{1}{L}\partial_a t_a - \frac{t_a}{L^2}\partial_a L ,
\end{align}
\begin{align}
    \partial_a L = 2c\,\partial_a c + 2d\,\partial_a d = 2cb + 2d[b]_\times n ,
\end{align}
\begin{align}
    \partial_a t_a = ([b]_\times n)\,b^T - b\,([b]_\times n)^T - c[b]_\times\!\left(\frac{1}{d}[b]_\times + \frac{s}{d^2}([b]_\times n)^T\right) .
\end{align}
The $ab$ block is
\begin{align}
    \partial_a\partial_b\theta = \frac{1}{L}\partial_a t_b - \frac{t_b}{L^2}\partial_a L ,
\end{align}
\begin{align}
    \partial_a t_b = -([a]_\times n)\,b^T + c[n]_\times + c[a]_\times\!\left(\frac{1}{d}[b]_\times + \frac{s}{d^2}([b]_\times n)^T\right) - a\,([b]_\times n)^T - d\, I_{3\times3} ,
\end{align}
\begin{align}
    \partial_b\partial_a\theta = (\partial_a\partial_b\theta)^T .
\end{align}
The $bb$ block is
\begin{align}
    \partial_b\partial_b\theta = \frac{1}{L}\partial_b t_b - \frac{t_b}{L^2}\partial_b L ,
\end{align}
\begin{align}
    \partial_b L = 2c\,\partial_b c + 2d\,\partial_b d = 2ca - 2d[a]_\times n ,
\end{align}
\begin{align}
    \partial_b t_b = a\,([a]_\times n)^T - ([a]_\times n)\,a^T - c[a]_\times\!\left(\frac{1}{d}[a]_\times + \frac{s}{d^2}([a]_\times n)^T\right) .
\end{align}

The solver does not evaluate this angle Hessian directly. To remain well-defined when weavers are nearly collinear at rest, where $\theta_0\to0$ or $\pi$ are the singular points of the angle-factored gradient, the implementation penalizes a smooth cosine measure instead, $\Pi=\frac{1}{2}k_{3spr}(\cos\theta-\cos\theta_0)^2$ with $\cos\theta=(a^Tb)/(\|a\|\|b\|)$, whose gradient and Hessian carry no $1/\|a\times b\|$ term and stay finite through the flat state. The angle gradient above is still used by the linear element matrices and by the energy maps, where it agrees with the compact equivalent form coded in \texttt{computeKlocal\_3spr},
\begin{align}
    \partial_a\theta = \frac{1}{d}\!\left(\frac{c}{\|a\|^2}\,a - b\right) , \qquad
    \partial_b\theta = \frac{1}{d}\!\left(\frac{c}{\|b\|^2}\,b - a\right) ,
\end{align}
which is algebraically identical to $(c\,\partial_a d - d\,\partial_a c)/L$ above.

\subsubsection*{A.1.4 Four-node springs}

The four-node spring penalizes the dihedral angle $\theta$ between the two panels meeting along a shared edge, supplying the inter-weaver out-of-plane bending and shear stiffness. This is the rotational hinge element of the bar-and-hinge origami literature \citep{Schenk2011,Liu2017,filipov2017bar}. The derivation mirrors the three-node spring, but the angle is now measured between two plane normals rather than two edges, so an extra chain step through the normal vectors is required. The energy and its first two derivatives take the same form as before,
\begin{align}
    \Pi_{4spr} = \frac{1}{2}k_{4spr}(\theta-\theta_0)^2 \in \mathbb{R}^{1\times1} ,
\end{align}
\begin{align}
    \partial \Pi_{4spr} = k_{4spr}(\theta-\theta_0)\,\partial\theta \in \mathbb{R}^{12\times1} ,
\end{align}
\begin{align}
    \partial^2 \Pi_{4spr} = k_{4spr}(\partial\theta)(\partial\theta)^T + k_{4spr}(\theta-\theta_0)\,\partial^2\theta \in \mathbb{R}^{12\times12} ,
\end{align}
where, as in the three-node case, the material term carries $(\partial\theta)(\partial\theta)^T$ and the geometric term carries $(\theta-\theta_0)\partial^2\theta$.

The reference angle is the angle between the two reference unit normals $\underline{n}_1$ and $\underline{n}_2$, and the current angle is built the same way from the deformed normals through $\operatorname{atan2}$, with the same robustness rationale as the three-node spring,
\begin{align}
    \theta_0 = \operatorname{atan2}\!\left(\|\underline{n}_1 \times \underline{n}_2\|,\ \underline{n}_1 \cdot \underline{n}_2\right) , \qquad
    \theta = \operatorname{atan2}(d,c) .
\end{align}
The four edge vectors that span the two panels are
\begin{align}
    \underline{v}_1 = \underline{x}_1 - \underline{x}_2 , \quad
    \underline{v}_2 = \underline{x}_3 - \underline{x}_2 , \quad
    \underline{v}_3 = \underline{x}_2 - \underline{x}_3 , \quad
    \underline{v}_4 = \underline{x}_4 - \underline{x}_3 ,
\end{align}
and the unit panel normals are the normalized cross products of the spanning edges of each panel,
\begin{align}
    \underline{n}_1 = \frac{\underline{v}_1 \times \underline{v}_2}{\|\underline{v}_1 \times \underline{v}_2\|} , \qquad
    \underline{n}_2 = \frac{\underline{v}_3 \times \underline{v}_4}{\|\underline{v}_3 \times \underline{v}_4\|} .
\end{align}
Their deformed counterparts use the displaced edges $\underline{v}_i'$,
\begin{align}
    \underline{v}_1' = (\underline{x}_1+\underline{u}_1) - (\underline{x}_2+\underline{u}_2) , \qquad
    \underline{v}_2' = (\underline{x}_3+\underline{u}_3) - (\underline{x}_2+\underline{u}_2) ,
\end{align}
\begin{align}
    \underline{v}_3' = (\underline{x}_2+\underline{u}_2) - (\underline{x}_3+\underline{u}_3) , \qquad
    \underline{v}_4' = (\underline{x}_4+\underline{u}_4) - (\underline{x}_3+\underline{u}_3) ,
\end{align}
giving the raw and normalized deformed normals
\begin{align}
    \tilde{a} = \underline{v}_1' \times \underline{v}_2' , \qquad
    \tilde{b} = \underline{v}_3' \times \underline{v}_4' ,
\end{align}
\begin{align}
    a = \frac{\tilde{a}}{\|\tilde{a}\|} , \qquad
    b = \frac{\tilde{b}}{\|\tilde{b}\|} .
\end{align}
From here the inner and cross product scalars are identical in form to the three-node spring, but with $a$ and $b$ now the unit normals,
\begin{align}
    c = a^T b , \qquad s = a \times b , \qquad d = \|s\| .
\end{align}

Because the angle now depends on $a$ and $b$, which depend on the normals, the chain has an extra layer. In normal space the gradients are exactly as before,
\begin{align}
    \partial_a\theta = \frac{c\,\partial_a d - d\,\partial_a c}{L} , \qquad
    \partial_b\theta = \frac{c\,\partial_b d - d\,\partial_b c}{L} , \qquad
    L = c^2 + d^2 = \|a\|^2\|b\|^2 ,
\end{align}
\begin{align}
    \partial_a c = b , \qquad \partial_b c = a , \qquad
    \partial d = n^T \partial s , \qquad n = \frac{s}{\|s\|} ,
\end{align}
\begin{align}
    s = a \times b = [a]_\times b = -[b]_\times a , \qquad
    \partial_a s = -[b]_\times , \qquad \partial_b s = [a]_\times ,
\end{align}
\begin{align}
    \partial_a d = [b]_\times n , \qquad \partial_b d = -[a]_\times n .
\end{align}
The extra layer is the derivative of the normalization map $\tilde{a}\mapsto a=\tilde{a}/\|\tilde{a}\|$, which is the scaled orthogonal projector that removes the component along the unit normal,
\begin{align}
    P_a = \frac{1}{\|\tilde{a}\|}\left(I - aa^T\right) , \qquad
    P_b = \frac{1}{\|\tilde{b}\|}\left(I - bb^T\right) .
\end{align}
Composing $P_a$ and $P_b$ with the skew-matrix derivatives of the raw normals $\tilde a=\underline{v}_1'\times \underline{v}_2'$ and $\tilde b=\underline{v}_3'\times \underline{v}_4'$ with respect to each contributing edge gives the per-node normal sensitivities,
\begin{align}
    A_1 = -P_a [\underline{v}_2']_\times , \quad
    A_2 = P_a \left([\underline{v}_1']_\times + [\underline{v}_2']_\times\right) , \quad
    A_3 = -P_a [\underline{v}_1']_\times ,
\end{align}
\begin{align}
    B_2 = P_b \left(-[\underline{v}_4']_\times - [\underline{v}_3']_\times\right) , \quad
    B_3 = P_b \left([\underline{v}_3']_\times - [\underline{v}_4']_\times\right) , \quad
    B_4 = P_b [\underline{v}_3']_\times .
\end{align}
These assemble into the map $T$ from normal-space gradients to the twelve nodal degrees of freedom,
\begin{align}
    T =
    \begin{bmatrix}
    A_1 & A_2 & A_3 & 0_{3\times3} \\
    0_{3\times3} & B_2 & B_3 & B_4
    \end{bmatrix} ,
\end{align}
\begin{align}
    \partial \theta = T^T
    \begin{bmatrix}
    \partial_a\theta \\
    \partial_b\theta
    \end{bmatrix}
    \in \mathbb{R}^{12\times1} .
\end{align}

The nodal Hessian is formed by pushing the normal-space block Hessian $H_{ab}$ through the same map $T$,
\begin{align}
    H_{\underline{u}}\theta = T^T H_{ab}\, T \in \mathbb{R}^{12\times12} , \qquad
    H_{ab} =
    \begin{bmatrix}
    \partial_a\partial_a\theta & \partial_a\partial_b\theta \\
    \partial_b\partial_a\theta & \partial_b\partial_b\theta
    \end{bmatrix} .
\end{align}
This expression neglects the curvature of the normalization map, that is, the second derivative of the unit normals with respect to the nodal coordinates. Those terms are zero at the reference state but nonzero away from it, so the four-node spring tangent given here is the exact material term together with an approximate geometric term. The approximation is consistent at and near the reference configuration and was found adequate for the quasi-static solves in this work. It is not, however, the exact consistent Hessian of the dihedral energy at large rotations, and a fully consistent tangent would require differentiating $T$ itself.

The normal-space block numerators are
\begin{align}
    t_a = c[b]_\times n - d\,b , \qquad
    t_b = -c[a]_\times n - d\,a ,
\end{align}
and the blocks have the same structure as the three-node spring,
\begin{align}
    \partial_a\partial_a\theta = \frac{1}{L}\partial_a t_a - \frac{t_a}{L^2}\partial_a L , \qquad
    \partial_a L = 2cb + 2d[b]_\times n ,
\end{align}
\begin{align}
    \partial_a t_a = ([b]_\times n)\,b^T - b\,([b]_\times n)^T - c[b]_\times\!\left(\frac{1}{d}[b]_\times + \frac{s}{d^2}([b]_\times n)^T\right) ,
\end{align}
\begin{align}
    \partial_a\partial_b\theta = \frac{1}{L}\partial_a t_b - \frac{t_b}{L^2}\partial_a L ,
\end{align}
\begin{align}
    \partial_a t_b = -([a]_\times n)\,b^T + c[n]_\times + c[a]_\times\!\left(\frac{1}{d}[b]_\times + \frac{s}{d^2}([b]_\times n)^T\right) - a\,([b]_\times n)^T - d\, I_{3\times3} ,
\end{align}
\begin{align}
    \partial_b\partial_a\theta = (\partial_a\partial_b\theta)^T ,
\end{align}
\begin{align}
    \partial_b\partial_b\theta = \frac{1}{L}\partial_b t_b - \frac{t_b}{L^2}\partial_b L , \qquad
    \partial_b L = 2ca - 2d[a]_\times n ,
\end{align}
\begin{align}
    \partial_b t_b = a\,([a]_\times n)^T - ([a]_\times n)\,a^T - c[a]_\times\!\left(\frac{1}{d}[a]_\times + \frac{s}{d^2}([a]_\times n)^T\right) .
\end{align}

The derivation above takes the angle between the unit normals and threads the chain through the projectors $P_a$ and $P_b$. The code (\texttt{computeKlocal\_4spr} and \texttt{computeFK\_4spr}) instead follows the more compact Liu and Paulino convention \citeyearpar{Liu2017}. It works with the raw unnormalized normals $n_1=v_1\times v_2$ and $n_2=-(v_2\times v_3)$ about the shared axis at nodes $2$ and $3$, and absorbs the normalization analytically into the $c/\|n\|^2$ terms of the gradient,
\begin{align}
    A_1=\frac{1}{D}\!\left(\frac{C}{\|n_1\|^2}n_1-n_2\right) , \qquad
    A_2=\frac{1}{D}\!\left(\frac{C}{\|n_2\|^2}n_2-n_1\right) ,
\end{align}
\begin{align}
    \partial_{v_1}\theta=v_2\times A_1 , \qquad
    \partial_{v_2}\theta=-(v_1\times A_1)-(v_3\times A_2) , \qquad
    \partial_{v_3}\theta=v_2\times A_2 ,
\end{align}
with $C=n_1\cdot n_2$ and $D=\|n_1\times n_2\|$. Both routes return the same $\theta$ and the same gradient, since the dihedral angle is unchanged by normalizing the normals. The projector form is the cleaner one to present analytically, and the raw-normal form is the one to cite when describing the implementation. As with the three-node spring, the nonlinear solver assembles the smooth cosine energy $\frac{1}{2}k_{4spr}(\cos\theta-\cos\theta_0)^2$ for robustness at flat and folded rest states, with this angle Hessian used for the linear matrices and the energy maps.

\subsection*{A.2. Numerical computation of derivatives}

As an independent check on the analytical expressions above, each element's Hessian can also be approximated by finite differencing the energy directly. We adopt a step size $\Delta \ll 1$ and form the second derivatives as follows.

The diagonal entries use the standard three-point central stencil in a single degree of freedom $u_{ij}$, with all other coordinates held at their reference values $\underline{u}=0$. For $ij = mn$,
\begin{equation}
    \label{eq:finDiffDiag}
    \frac{\partial^2\Pi}{\partial u_{ij}^2} = \frac{1}{\Delta^2}\Big[ \Pi\big(x_{ij}+\Delta,\,\underline{x}_{\widehat{ij}}\big) - 2\,\Pi\big(x_{ij},\,\underline{x}_{\widehat{ij}}\big) + \Pi\big(x_{ij}-\Delta,\,\underline{x}_{\widehat{ij}}\big) \Big],
\end{equation}
where $\underline{x}_{\widehat{ij}}$ denotes all coordinates other than the one being perturbed. This stencil is the usual $O(\Delta^2)$ approximation to $\partial^2\Pi/\partial u_{ij}^2$.

The off-diagonal terms couple two distinct degrees of freedom and therefore use the four-point mixed central stencil, perturbing $u_{ij}$ and $u_{mn}$ simultaneously in all four sign combinations. For $ij \neq mn$,
\begin{equation}
    \label{eq:finDiffOffDiag}
    \frac{\partial^2\Pi}{\partial u_{ij}\,\partial u_{mn}} = \frac{1}{4\Delta^2}\Big[ \Pi(+\Delta,+\Delta) - \Pi(+\Delta,-\Delta) - \Pi(-\Delta,+\Delta) + \Pi(-\Delta,-\Delta) \Big],
\end{equation}
where each $\Pi(\pm\Delta,\pm\Delta)$ denotes the energy with $x_{ij}$ and $x_{mn}$ perturbed by the indicated signs and all remaining coordinates held at reference. This symmetric pattern is the $O(\Delta^2)$ approximation to the mixed partial $\partial^2\Pi/\partial u_{ij}\partial u_{mn}$ and yields a symmetric numerical Hessian, as it must. In practice this finite-difference Hessian is computed once per element type at representative configurations and compared entry by entry against the analytical matrices of Section~A.1. For the Cartesian connector, bar, and three-node spring, whose analytical tangents are exact, the two agree to near machine precision at every configuration tested, with errors on the order of $10^{-9}$. For the four-node spring the same agreement holds at and near the reference configuration, where the neglected normalization-map curvature vanishes; away from the reference state the finite-difference Hessian departs from the analytical matrix by exactly that curvature term, consistent with the approximate geometric tangent noted in Section~A.1.4. These checks confirm the analytical derivations.
\FloatBarrier

\bibliographystyle{cas-model2-names}

\bibliography{references}
\end{document}